\documentclass[lettersize,journal]{IEEEtran}
\usepackage{amsmath,amsfonts}
\usepackage{algorithm}
\usepackage{algorithmicx}
\usepackage{algpseudocode}
\usepackage{array}
\usepackage{textcomp}
\usepackage{stfloats}
\usepackage{url}
\usepackage{verbatim}
\usepackage{graphicx}
\usepackage{cite}
\usepackage{csquotes}
\usepackage{rotating}
\usepackage{multirow}
\usepackage{times}
\usepackage{epsfig}
\usepackage{color}
\usepackage[normalem]{ulem}
\usepackage{booktabs}
\usepackage{subfig}

\usepackage{colortbl,xcolor}
\definecolor{Gray}{gray}{0.8}
\usepackage[colorlinks]{hyperref}
\usepackage[capitalize]{cleveref}
\crefname{section}{Sec.}{Secs.}
\Crefname{section}{Section}{Sections}
\Crefname{table}{Table}{Tables}
\crefname{table}{Tab.}{Tabs.}

\begin{document}

\title{MC-Blur: A Comprehensive Benchmark for Image Deblurring}


\author{Kaihao Zhang,~
        Tao Wang,~
        Wenhan Luo,~
        Wenqi Ren,~
        Björn Stenger,~
        Wei Liu, \textit{Fellow, IEEE}~
        \\
        Hongdong Li,~
        Ming-Hsuan Yang, \textit{Fellow, IEEE}
        \\
\IEEEcompsocitemizethanks{\IEEEcompsocthanksitem K. Zhang and H. Li are with the College of Engineering and Computer Science, Australian National University, Canberra, Australia (e-mail: \{kaihao.zhang, hongdong.li\}@anu.edu.au).
\IEEEcompsocthanksitem T. Wang is with the State Key Laboratory for Novel Software Technology, Nanjing University, Nanjing, China (e-mail: taowangzj@gmail.com).
\IEEEcompsocthanksitem W. Luo and W. Ren are with Sun Yat-sen University, Guangzhou, China, 
(e-mail: \{whluo.china, rwq.renwenqi\}@gmail.com).
\IEEEcompsocthanksitem B. Stenger is with Rakuten Institute of Technology, Setagaya, Tokyo 
(e-mail: bjorn.stenger@gmail.com).
\IEEEcompsocthanksitem W. Liu is with Tencent Data Platform, Shenzhen, China 
(e-mail: wl2223@columbia.edu).
\IEEEcompsocthanksitem M. Yang is with School of Engineering, University of California at Merced, Merced, CA, USA 
(e-mail: mhyang@ucmerced.edu).}

}

\markboth{Journal of \LaTeX\ Class Files,~Vol.~14, No.~8, August~2021}%
{Shell \MakeLowercase{\textit{et al.}}: A Sample Article Using IEEEtran.cls for IEEE Journals}


\maketitle

\begin{abstract}
Blur artifacts can seriously degrade the visual quality of images, and numerous deblurring methods have been proposed for specific scenarios. However, in most real-world images, blur is caused by different factors, \textit{e.g.}, motion, and defocus. In this paper, we address how other deblurring methods perform in the case of multiple types of blur. For in-depth performance evaluation, we construct a new large-scale multi-cause image deblurring dataset (MC-Blur), including real-world and synthesized blurry images with different blur factors. The images in the proposed MC-Blur dataset are collected using other techniques: averaging sharp images captured by a $1000$-fps high-speed camera, convolving Ultra-High-Definition (UHD) sharp images with large-size kernels, adding defocus to images, and real-world blurry images captured by various camera models. Based on the MC-Blur dataset, we conduct extensive benchmarking studies to compare SOTA methods in different scenarios, analyze their efficiency, and investigate the buildataset's capacity. These benchmarking results provide a comprehensive overview of the advantages and limitations of current deblurring methods, revealing our dataset's advances.
The dataset is available to the public at \href{http://github.com/HDCVLab/MC-Blur-Dataset}{https://github.com/HDCVLab/MC-Blur-Dataset}.
\end{abstract}


\section{Introduction}
\label{introduction}
Image deblurring is an important problem in computer vision and image processing, which aims to restore a sharp image from an observed blurry input \cite{zhang2022deep}. 
Deblurring has been widely used in medical image analysis, computational photography, and video enhancement applications. 
Conventional methods formulate the task as an inverse filtering problem using the uniform blur model:
\begin{equation}
\label{eq:formula_blur}
    I_B = I_S * K + \sigma_N,
\end{equation}
where $I_B$ is the observed blurry image, $I_S$ is the latent sharp image, $K$ is the unknown blur kernel, $\sigma_N$ is the additive noise, and $*$ is the convolution operation. 
Image deblurring is a well-known ill-posed problem, and numerous priors, such as natural image statistics, have been employed to constrain the solution space. 
However, estimating $I_S$ based on this formulation typically involves iterative and time-consuming estimation processes.

\begin{figure*}[tb]
  \centering
  \subfloat[Real high-FPS motion-blurred images from the RHM subset of the proposed MC-Blur dataset.]{
    \label{fig:results_avg}
    \includegraphics[width= \linewidth]{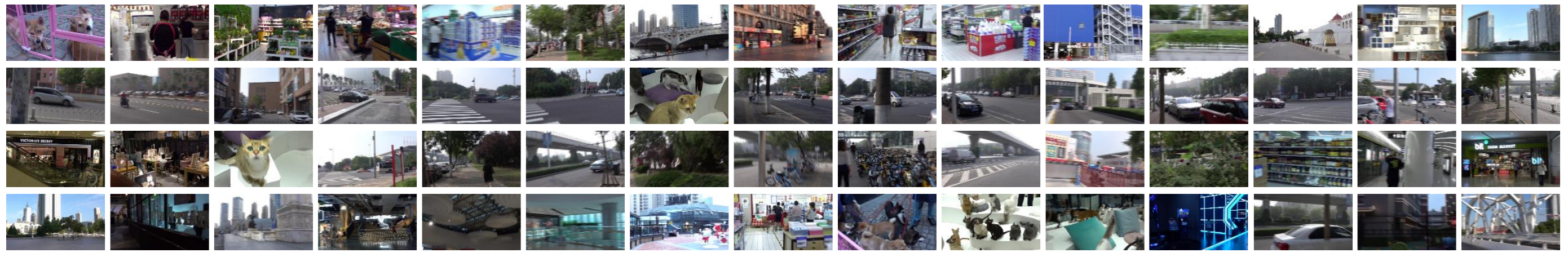}}\\
  \subfloat[Sample images of UHD motion-blurred images using large kernels from the proposed UHDM subset of our MC-Blur dataset.]{
    \label{fig:results_conv}
    \includegraphics[width=\linewidth]{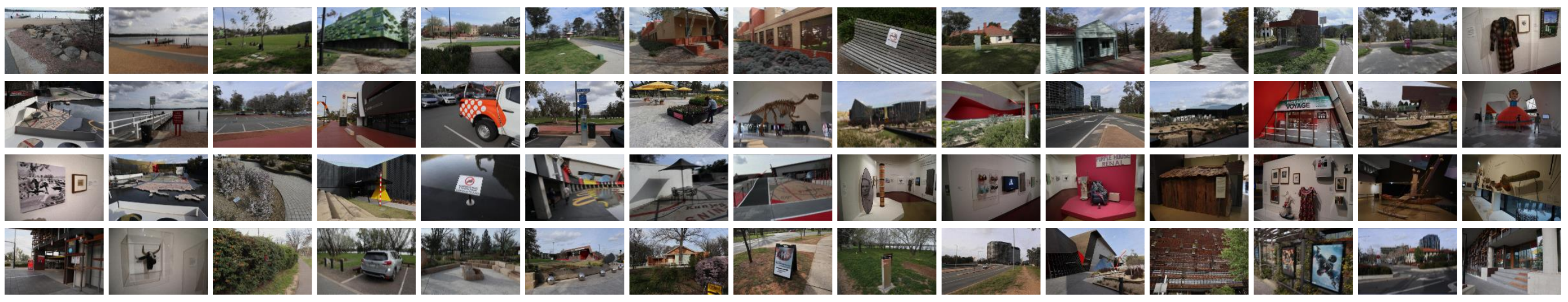}}\\
    \subfloat[Defocus blurry images from the LSD subset of our MC-Blur dataset.]{
    \label{fig:example_defocus}
    \includegraphics[width=\linewidth]{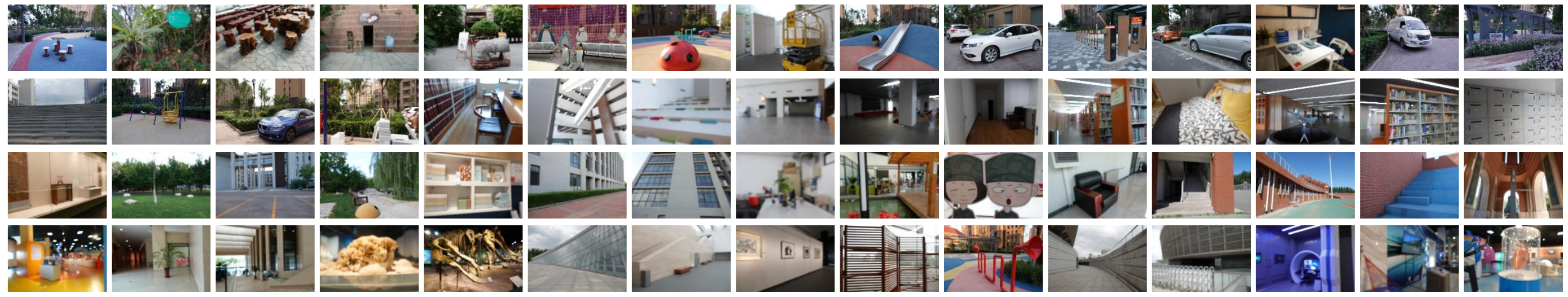}}\\
   \subfloat[The real-world blurry images from the RMBQ subset of our MC-Blur dataset.]{
    \label{fig:results_real}
    \includegraphics[width=\linewidth]{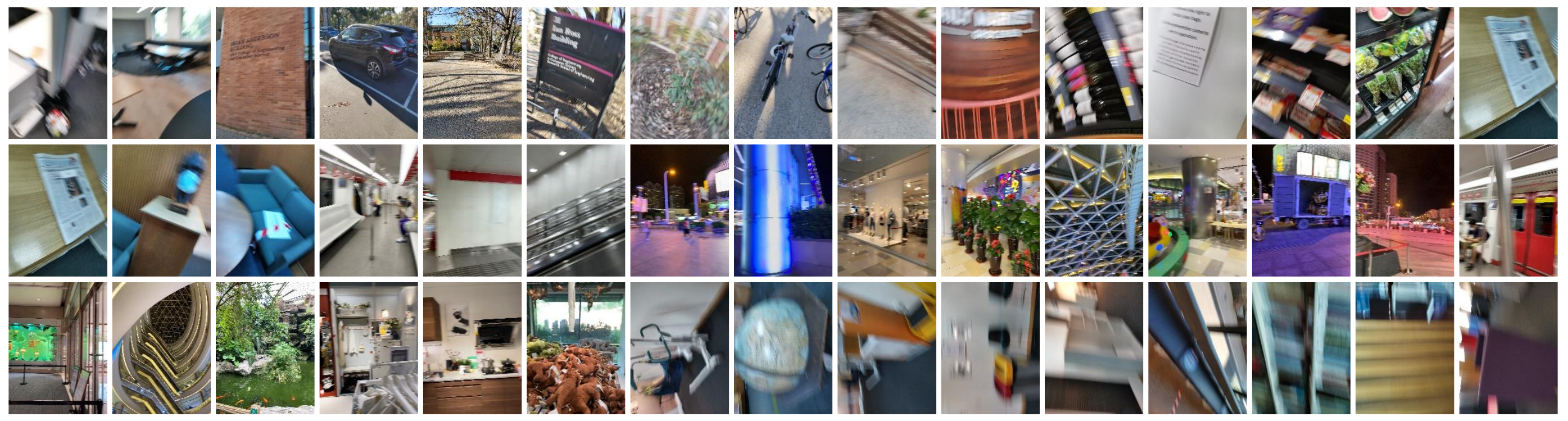}}
   \caption{\textbf{Sample blurry images of different subsets from our MC-Blur dataset.} From top to bottom are RHM, UHDM, LSD, and RMBQ subsets from our proposed MC-Blur dataset. The dataset comprises images sourced from over $1,000$ diverse scenes, including buildings, cityscapes, vehicles, natural landscapes, people, animals, and sculptures.} 
  \label{fig:mcid_dataset} 
\end{figure*}

Numerous deep models have recently been applied to image deblurring within the supervised learning framework.
These models require a large number of paired sharp and blurry images to train networks in an end-to-end manner. 
Several datasets have been created by averaging continuous frames, convolving with blur kernels, or directly taking photos with two cameras with different shutter durations to obtain pairs of images.
Although these datasets have advanced the deep deblurring models, there remain unaddressed issues with these datasets:
1) As shown in~\cite{Nah_2019_CVPR_Workshops_REDS}, averaging sharp images of low frame rate to synthesize blurry images can cause unnatural blur. 
For datasets that include motion blur, images are usually generated by averaging continuous frames captured with a relatively \textbf{slow} and \textbf{fixed} shutter speed (\textit{e.g.}, the GoPro dataset (240 fps)), or from images in interpolated high fps videos (\textit{e.g.}, the REDS dataset~\cite{Nah_2019_CVPR_Workshops_REDS}); 
2) For datasets containing non-uniform blur, \textit{e.g.}, the dataset by K\"ohler et al.~\cite{kohler2012recording}, the number of images is insufficient for training deep networks, images are not of high definition, and the kernel size is relatively small. 
With an increasing number of devices being able to capture Ultra-High-Definition (UHD) images, existing datasets are not suitable for training models capable of handling such images;
3) Datasets of real-world blurry images typically require additional processing steps such as accurate alignment~\cite{rim2020real};
4) While defocus is a common cause of blurry images, few datasets are explicitly developed for this type of blur. In addition, existing ones like~\cite{abuolaim2020defocus} are usually of small scale or lack images of heavy defocus blur, making them infeasible for studying heavy defocus deblurring.

To overcome these limitations, we construct a comprehensive and large-scale multi-cause dataset, including blurry images caused by multiple factors, named  {\it MC-Blur} dataset (See Fig. \ref{fig:mcid_dataset}). 
This dataset is composed of four subsets. 
The first one, Real High-fps based Motion-blurred subset (RHM), includes images averaged from sharp frames to synthesize motion blur.  
Unlike existing datasets, sharp frames in RHM are captured with various ultra-high-speed cameras (iPhone, Samsung, Sony, etc.) at different frame rates ($250$, $500$, and $1000$ fps). 
With different types of devices and frame rates, this subset mimics various motion blur in the real world. 
The second one, the large-kernel UHD Motion-blurred subset (UHDM), contains motion blur based on convolving sharp images with blur kernels.
Due to the increasing number of high-definition cameras, we capture many UHD images at 4K+ resolution. These UHD images are convolved with large blur kernels. 
The third subset, LSD (Large-Scale Defocus), is specific to defocus blur. 
We capture images with various heavy defocus effects by manually changing the focus setting. 
The fourth one, the Real Mixed Blurry Qualitative subset (RMBQ), comprises real-world blurry images captured by different types of devices, \textit{e.g.}, mobile cameras. 
While no sharp images are available as ground truth, this subset is included for qualitative performance evaluation in real-world scenarios. 

The main contributions of this paper are summarized as follows.
\begin{itemize}
\item
We build a large-scale and comprehensive dataset (MC-Blur) of images containing blur artifacts due to multiple causes:
\begin{itemize}

\item
The RHM subset provides motion-blurred images synthesized from real, higher-and-unfixed fps video frames without artificially interpolated technologies. Experimental studies demonstrate its superior generalization potential compared with the widely employed ones.
\item
The UHDM subset is the first large-scale UHD image deblurring dataset expected to drive future research regarding the problem of single UHD image deblurring.  
\item
The LSD subset is the largest defocus blurry dataset, providing blurry images from the real world of heavier defocus artifacts compared with existing ones. This large amount of images of more serious artifacts will benefit our community in exploring the problem of heavy defocus deblurring.
\item
The RMBQ subset provides large-scale, real blurry images captured by various mobile devices, serving as a credible testing bench for the qualitative study of future research in terms of real-world scenarios.  
\end{itemize}

\item
We carry out extensive benchmarking analysis of recent state-of-the-art image deblurring methods on our MC-Blur dataset. The benchmark study, including evaluating main-stream image deblurring methods, efficiency analysis, and effectiveness of cross-dataset learning, provides a comprehensive understanding of the SOTA methods in various scenarios.

\end{itemize}

The subsequent sections are structured as follows. Section~\ref{related_work} provides an overview of related works on image deblurring datasets and methods. In Section~\ref{benchmark_dataset}, the specifics of the proposed MC-Blur dataset are presented.
Section~\ref{Experiments} showcases the benchmarking results of existing deblurring approaches on the aforementioned MC-Blur dataset. Finally, Section~\ref{Conclusion} summarizes the findings and conclusions of this study.


\section{Related Work}
\label{related_work}
In this section, we provide an overview of the datasets commonly employed for the image deburring task. Subsequently, we review existing image deblurring methods in the literature.
\subsection{Image Deblurring Datasets}
Several datasets have been developed for advances in image deblurring~\cite{levin2009understanding,sun2012super,kohler2012recording,lai2016comparative,nah2017deep,shen2019human,jiang2020learning,su2017deep,hradivs2015convolutional,shen2018deep,zhou2019davanet}. 
For example, motion blur is simulated by convolving images with uniform blur kernels by Levin \textit{et al.} ~\cite{levin2009understanding} and Sun \textit{et al.} ~\cite{sun2012super}, or with non-uniform kernels by K\"ohler \textit{et al.} ~\cite{kohler2012recording}.
In~\cite{lai2016comparative}, Lai \textit{et al.} introduce a dataset including images of both real-world and synthetic blur (uniform blur kernels).
However, the size of this dataset is still relatively small. 
Even when synthesizing blurry images via realistic blur kernels, the scale of the datasets mentioned above is small, making them difficult for deep learning-based deblurring methods. 
Furthermore, the blur kernels are relatively small, making them less effective for deblurring Ultra-High-Definition (UHD) images.

Considering that images are captured within the duration of camera exposure, blurry images can be modeled by the integration of neighboring frames~\cite{hirsch2011fast},
\begin{equation}
\label{blurr_process1}
I_B  = g \left(\frac{1}{T}{\int_{t=0}^{T}I_{S(t)}\mathrm{d}t} \right), 
\end{equation}
where $T$ is the exposure time period and $g(\cdot)$ is the Camera Response Function (CRF). 
To model this process~\cite{hirsch2011fast} and alleviate the problem of alignment \cite{rim2020real}, several deblurring datasets have been created based on the discrete formulation,
\begin{equation}
\label{blurr_process2}
I_B \simeq g \left( \frac{1}{M}\sum_{t=0}^{M-1}I_{S[t]} \right),
\end{equation}
where $M$ is the number of frames. 

In particular, the GoPro dataset~\cite{nah2017deep} has been widely used for training deep models. 
Its sharp images are captured by a GoPro Hero4Black camera with a shutter speed of $240$ fps. 
Blurry images are generated by averaging continuous sharp frames over a time window. 
Similarly, based on this method, the HIDE~\cite{shen2019human} and REDS~\cite{Nah_2019_CVPR_Workshops_REDS} datasets are created.

Rim \textit{et al.} ~\cite{rim2020real} develop a dataset containing real blurry images and the corresponding sharp images. 
Two different cameras take image pairs with varying times of exposure.
While the blur is realistic, this work requires an additional image alignment step to generate image pairs, which causes the problem of imprecise alignment. 
In addition, this dataset also lacks defocus blurry images or UHD images, which is of great interest for real-world scenarios. 
On the other hand, Abuolaim and Brown~\cite{abuolaim2020defocus} capture $500$ images with defocus blur, but this number is small compared to the recent large-scale deblurring datasets. Moreover, the extent of defocus blur on their blurry images is relatively slight.
The details of the existing representative datasets and the proposed MC-Blur dataset are listed in Table~\ref{table_dataset}.

\subsection{Deblurring Methods}

In the computer vision community, the problem of image deblurring has garnered significant attention owing to its inherently ill-posed nature. Consequently, numerous image deblurring methods have emerged in the literature. Broadly speaking, these methods can be categorized into two main groups: 1) conventional deblurring methods and 2) deep learning-based deblurring methods. We introduce these methods in the following.

\textbf{Conventional Deblurring Methods}. Early deblurring methods refer to traditional approaches that address the image deblurring problem by incorporating constraints on blur kernels or latent images~\cite{bai2019single,wen2020simple,luo2021blind}. Consequently, lots of effective priors have been proposed, including the sparse gradients distribution model~\cite{pan2014deblurring}, dark channel prior~\cite{pan2017deblurring}, Local Maximum Gradient prior~\cite{chen2019blind}, structure prior~\cite{bai2019single}, superpixel segmentation prior~\cite{luo2021blind}, and more. However, most image deblurring methods primarily focus on mitigating blur caused by camera movement. At the same time, real dynamic scenes involve additional complexities, such as camera movement, object movement (rigid or non-rigid), and variations in scene depth. Consequently, these methods often encounter challenges when handling blur within dynamic scenes.


\textbf{Deep Deblurring Methods}. In recent years, numerous deep learning methods have been proposed to address various computer vision tasks \cite{wang2023self,li2018joint,li2018dynamic}, which also include single image deblurring \cite{sun2015learning,chakrabarti2016neural,nimisha2017blur,xu2017motion,jin2018learning}, and video deblurring \cite{hyun2017online,aittala2018burst,nah2019recurrent,wang2019edvr}.
Deep deblurring methods typically train neural networks in an end-to-end manner, using blurry images as inputs and updating network parameters by comparing the outputs and the ground truth sharp images \cite{mustaniemi2019gyroscope,aljadaany2019douglas,kaufman2020deblurring,jiang2020learning}. 
The idea of multi-scale processing and an adversarial loss is used in~\cite{nah2017deep} for image deblurring.  
Kupyn \textit{et al.} ~\cite{kupyn2018deblurgan} adopt a conditional GAN for the deblurring task, resulting in the DeblurGAN model. 
An enhancement in terms of accuracy and speed is introduced with DeblurGAN-v2~\cite{kupyn2019deblurgan}. 
In \cite{chakrabarti2016neural}, patch-level deblurring is carried out to obtain an initial global estimation.
The blur kernel is estimated, and the final result is obtained via deconvolution.
Motivated by Spatial Pyramid Matching, a multi-patch scheme is applied to learn hierarchical representations in~\cite{Zhang_2019_CVPR}.
A CNN is combined with an RNN in~\cite{zhang2018dynamic} to deblur images of spatial-variant blur in dynamic scenes.
Similarly, an LSTM with a CNN is employed in~\cite{tao2018scale}.
More recently, Zamir \textit{et al.} \cite{zamir2021multi} introduce a multi-stage architecture that gradually learns restoration functions for the degraded inputs, thus decomposing the overarching recovery process into a series of more manageable steps.
They \cite{zamir2021restormer} also develop an efficient Transformer-based model designed to address the limitations of convolutional neural networks (CNNs) and enable effective image restoration.
In addition, new architectures including skip connections \cite{gao2019dynamic,ren2019neural,zhou2019spatio,pan2020cascaded,zhang2018adversarial,niu2021blind,niu2021deep}, reblurring networks \cite{zhang2020deblurring,chen2018reblur2deblur,rim2022realistic}, unsupervised/self-supervised learning \cite{madam2018unsupervised,lu2019unsupervised,li2023self}, attention blocks \cite{purohit2019region,suin2020spatially,park2020multi,zhang2021multi,sun2022event,kong2023efficient,dudhane2023burstormer,park2023all,li2021arvo}, NAS \cite{hu2021pyramid} and Nerf \cite{wang2023bad}, are developed to improve the image deblurring performance.

\begin{figure*}[t] 
  \centering
  {\includegraphics[width=0.99\linewidth]{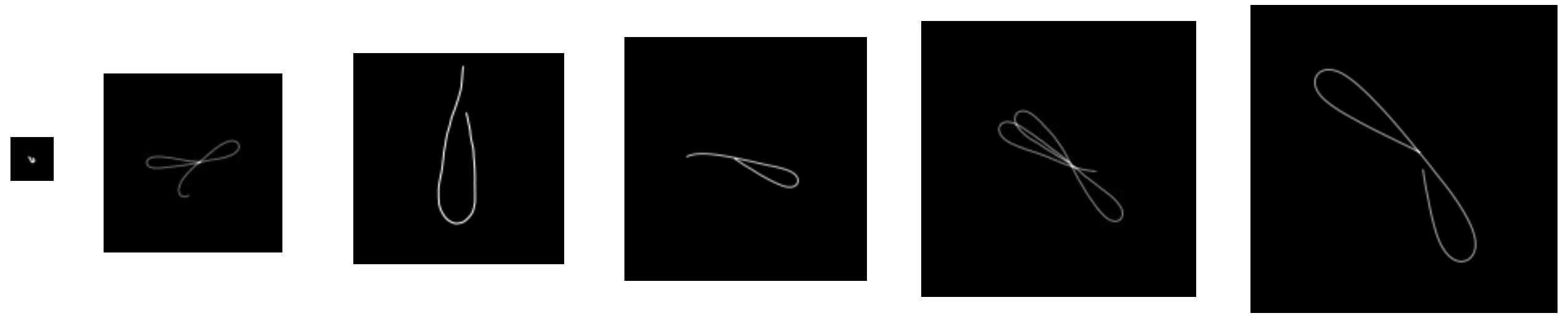}}
  \caption{\textbf{Blur kernels with different sizes.} From left to right, the sizes of blur kernels are $15\times15$, $111\times111$, $131\times131$, $151\times151$, $171\times171$ and $191\times191$.  We use larger blur kernels to create a UHDM subset. }
  \label{fig:blur_kernel} 
\end{figure*}

\begin{table*}[t]
\small
  \centering 
    \caption{Representative benchmark datasets for evaluating single image deblurring algorithms. 
    }
    	\scalebox{0.99}{\begin{tabular}{l  c c c c c c}
    \toprule
     \cellcolor{Gray}  Dataset &   \cellcolor{Gray}  Sharp  &   \cellcolor{Gray} Blurred &  \cellcolor{Gray}  Motion &  \cellcolor{Gray}  Defocus &  \cellcolor{Gray}  Real &   \cellcolor{Gray} Aligned \\
    \hline
    Levin \textit{et al.} ~\cite{levin2009understanding}  & 4 & 32 & $\checkmark$ & $\times$ & $\times$ & $\checkmark$ \\
    Sun \textit{et al.} ~\cite{sun2012super}& 80 & 640 & $\checkmark$ & $\times$ & $\times$ & $\checkmark$ \\
    K\" ohler \textit{et al.} ~\cite{kohler2012recording} & 4 & 48 & $\checkmark$ & $\times$ & $\times$ & $\checkmark$ \\
    Lai \textit{et al.} ~\cite{lai2016comparative} & 108 & 300 & $\checkmark$ & $\times$ & $\checkmark$ & $\checkmark$ \\
    GoPro \cite{nah2017deep} & 3,214 & 3,214 & $\checkmark$ & $\times$ & $\times$ &  $\checkmark$ \\
    HIDE~\cite{shen2019human} & 8,422 & 8,422 & $\checkmark$ & $\times$ & $\times$ & $\checkmark$ \\ 
    Blur-DVS \cite{jiang2020learning} & 2,178 & 2,918 & $\checkmark$ & $\times$ & $\times$ & $\checkmark$ \\
    Abuolaim \textit{et al.} \cite{abuolaim2020defocus} & 500 & 500 & $\times$ & $\checkmark$ & $\checkmark$ & $\checkmark$ \\
    RealBlur \cite{rim2020real} & 9476 & 9476 & $\checkmark$ & $\times$ & $\checkmark$ & $\times$ \\
    \hline
    \textbf{RHM-250fps} & \textbf{25,000} & \textbf{25,000} & $\checkmark$ & $\times$ & $\times$ & $\checkmark$ \\
    \textbf{RHM-500fps} & \textbf{25,000} & \textbf{25,000} & $\checkmark$ & $\times$ & $\times$ & $\checkmark$ \\
    \textbf{RHM-1000fps} & \textbf{37,500} & \textbf{37,500} & $\checkmark$ & $\times$ & $\times$ & $\checkmark$ \\
    \textbf{UHDM} & \textbf{2,000} & \textbf{10,000} & $\checkmark$ & $\times$ & $\times$ & $\checkmark$ \\
    \textbf{LSD} & \textbf{2,800} & \textbf{2,800} & $\times$ & $\checkmark$ & $\checkmark$ &  $\checkmark$ \\
    \textbf{RMBQ} & \textbf{-} & \textbf{10,000} & $\checkmark$ & $\checkmark$ & $\times$ & $\checkmark$ \\
    \hline
    \textbf{MC-Blur dataset} & \textbf{92,300} & \textbf{110,300} & $\checkmark$ & $\checkmark$ & $\checkmark$ & $\checkmark$ \\
    \bottomrule
    \end{tabular}}%
    \label{table_dataset}
\end{table*}%

\section{MC-Blur Dataset}
\label{benchmark_dataset}

The advances in learning-based methods for image deblurring rely heavily on the quality and scale of datasets. 
To benchmark the state-of-the-art image deblurring methods in various conditions, we construct the large-scale multi-cause (MC-Blur) dataset. 
It consists of four blur types: uniform blur, motion blur by averaging continuous frames, heavy defocus blur, and real-world blur. 
In addition, the MC-Blur dataset includes many images captured during day and night time.
We collect these images from over 1000 diverse scenes such as buildings, city scenes, vehicles, natural landscapes, people, animals, and sculptures.
The four subsets are introduced in the following.

\noindent \textbf{RHM Subset.}
Averaging continuous frames within a time window to generate motion-blurred images is a common practice for synthesizing images with motion blur. 
For example, in the GoPro dataset, images captured at $240$ fps are used to produce blurry images~\cite{nah2017deep}.
However, if the frame rate of the images to be averaged is not sufficiently high, the synthesized motion blur can be unnatural~\cite{Nah_2019_CVPR_Workshops_REDS}. 
As such, Nah \textit{et al.} ~\cite{Nah_2019_CVPR_Workshops_REDS} record videos at $120$ fps and interpolate them to $1920$ fps by CNNs.
We capture sharp images using high-frame-rate cameras to remove this potential error source to create the real high fps-based motion-blurred dataset, RHM.
Images in RHM are captured in three settings. 
The first setting corresponds to the highest fps, up to $1,000$ fps. 
The sharp videos are recorded using a Sony RX10 camera. This subset contains $30,000$ images for training and $7,500$ for testing.
The sharp images in the second setting are captured with the same camera at $500$ fps. 
Ultra-high-speed (UHS) cameras usually adopt the MEMC (motion estimation and compensation) module, \textit{i.e.}, frame interpolation, to increase the frame rate. In this paper, all UHS frames are captured without using MEMC.
The training and testing sets contain $20,000$ and $5,000$ images, respectively. 
The third setting corresponds to images captured at $250$ fps with mobile devices such as iPhone, Huawei, and Sony RX10 cameras. 
For training and testing, this set contains $20,000$ and $5,000$ images, respectively. 
All images are resized via bicubic downsampling to reduce noise. 
The image resolutions in this set are $960 \times 540$ and $640 \times 360$ pixels.

\noindent \textbf{UHDM Subset.}
Another way to synthesize degraded images caused by motion blur is to convolve images with kernels. 
Existing datasets based on this approach use low-resolution images or small blur kernels. 
For example, when the image resolution is lower than $720 \times 720$ pixels, the size of the blur kernels is usually set within the range from $15 \times15$ to $27 \times 27$ pixels.
We note that deblurring 4K+ images requires restoration with more details, which is challenging if the models are trained with low-resolution images.
To address this critical concern and ensure our dataset mirrors real-world scenarios, we capture sharp images of 4K-6K resolutions to create the large-kernel UHD motion-blurred set, UHDM.
In our quest for realism, we employ blur kernels of varying sizes — $111 \times 111$, $131 \times 131$, $151 \times 151$, $171 \times 171$, and $191 \times 191$ — to convolve with the sharp images. This meticulous attention to diverse kernel sizes imbues our dataset's heightened sense of authenticity.
The training and testing sets contain $8,000$ and $2,000$ images, respectively. 
Blur kernels are generated via 3D camera trajectories \cite{boracchi2012modeling}.
The blur kernels with different sizes are shown in Fig. \ref{fig:blur_kernel}.

\begin{table*}[t]
  \centering 
    \caption{Performance evaluation of deep image deblurring methods on the proposed RHM set. PSNR and SSIM values are reported.}
    {
    \begin{tabular}{c  c c c c }
    \toprule
    \cellcolor{Gray} Method  &  \cellcolor{Gray} DeepDeblur \cite{nah2017deep} & \cellcolor{Gray} DeblurGAN \cite{kupyn2018deblurgan} & \cellcolor{Gray} SRN \cite{tao2018scale} & \cellcolor{Gray} DeblurGAN-v2 \cite{kupyn2019deblurgan}  \\
    \hline 
    250fps & 30.38/0.8766  & 24.89/0.6364 & 30.57/0.8799  & 26.99/0.8061    \\
    500fps & 31.08/0.8974  & 24.66/0.6748 & 31.54/0.9051 & 27.67/0.8320   \\
    1000fps & 32.41/0.8966 & 25.20/0.6535 & 32.69/0.9016 & 29.81/0.8461   \\
    \hline 
    \cellcolor{Gray} DMPHN \cite{Zhang_2019_CVPR}  &  \cellcolor{Gray} DBGAN \cite{zhang2020deblurring} & \cellcolor{Gray} MPRNet \cite{zamir2021multi} & \cellcolor{Gray} Restormer \cite{zamir2021restormer} & \cellcolor{Gray} MIMO-UNet \cite{cho2021rethinking} \\
    \hline 
    30.42/0.8768 & 27.89/0.8191 & \underline{\textit{31.52/0.9239}} 
& \textbf{32.02/0.9285} & 31.42/0.9211  \\
     31.43/0.9018 & 28.36/0.8388 & \underline{\textit{32.08/0.9300}} & 30.98/0.9160 & \textbf{32.89/0.9398}  \\
      32.41/0.9096 & 29.66/0.8318 & \underline{\textit{33.36/0.9332}} & 32.77/0/9264 & \textbf{33.75/0.9360}  \\
    \bottomrule
    \end{tabular}}
    \label{table:results_avg}
\end{table*}

\vspace{2mm}
\noindent \textbf{LSD Subset.}
A few datasets on defocus image deblurring have recently been developed. 
To create the Dual-Pixel \cite{abuolaim2020defocus} dataset, Abuolaim \textit{et al.} capture pairs of images of the same static scene at two aperture sizes via a Canon EOS 5D Mark IV DSLR camera. 
Focus distance and focal length differ across captured pairs to capture various defocus blurry images. 
However, this dataset is mainly designed for the dual-pixel problem. 
As a result, it provides only $500$ pairs of blurry images and their corresponding all-in-focus images, and the scale is small for approaches based on deep learning models. 
Meanwhile, these blurry images contain large sharp patches, and their extent of defocus blur is relatively slight. 
Given existing deep deblurring networks mainly take patches cropped from blurry images as input, the role of these sharp areas is less important during the training stage.

We collect a Large-Scale heavy Defocus blurred set, LSD, consisting of $2,250$ image pairs of sharp images and blurry images with the defocus effect in the training set, and $550$ image pairs for testing. 
The image resolution is at least $3600 \times 2400$ pixels. For the convenience of training, we crop sixteen patches ($900 \times 600$) without overlap from each image during the training and testing stages. 
Unlike the DPDD dataset, whose blurry images (about $1680 \times 1120$) contain large sharp regions, images in our LSD are completely defocused without any sharp region.
We manually control the focus to obtain the heavily blurry images and their corresponding sharp ones.

\begin{figure*}[h]
  \centering
  \subfloat[Input]{
    \includegraphics[width= 0.33\linewidth]{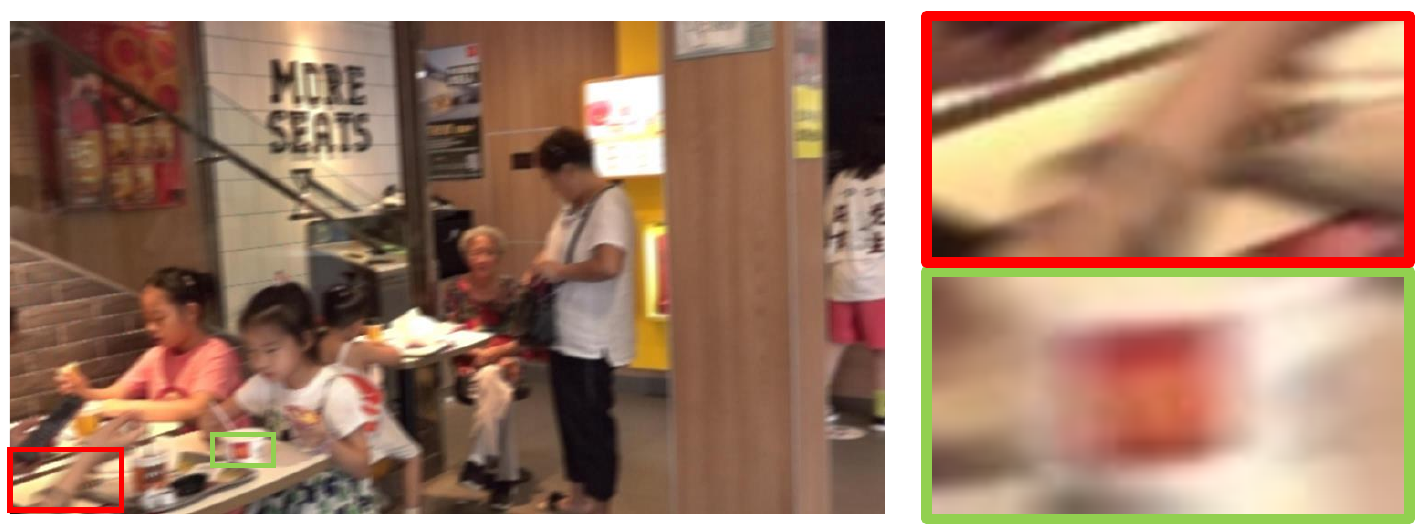}}
  \subfloat[DeepDeblur \cite{nah2017deep}]{
    \includegraphics[width=0.33\linewidth]{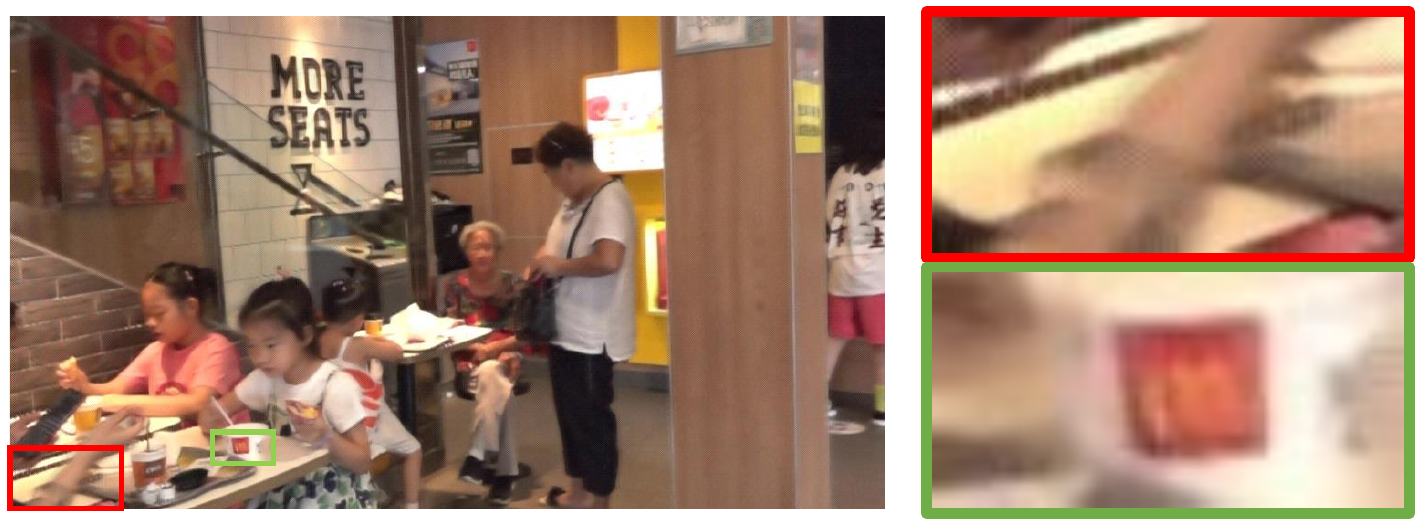}}
    \subfloat[DeblurGAN-v2 \cite{kupyn2019deblurgan}]{
    \includegraphics[width=0.33\linewidth]{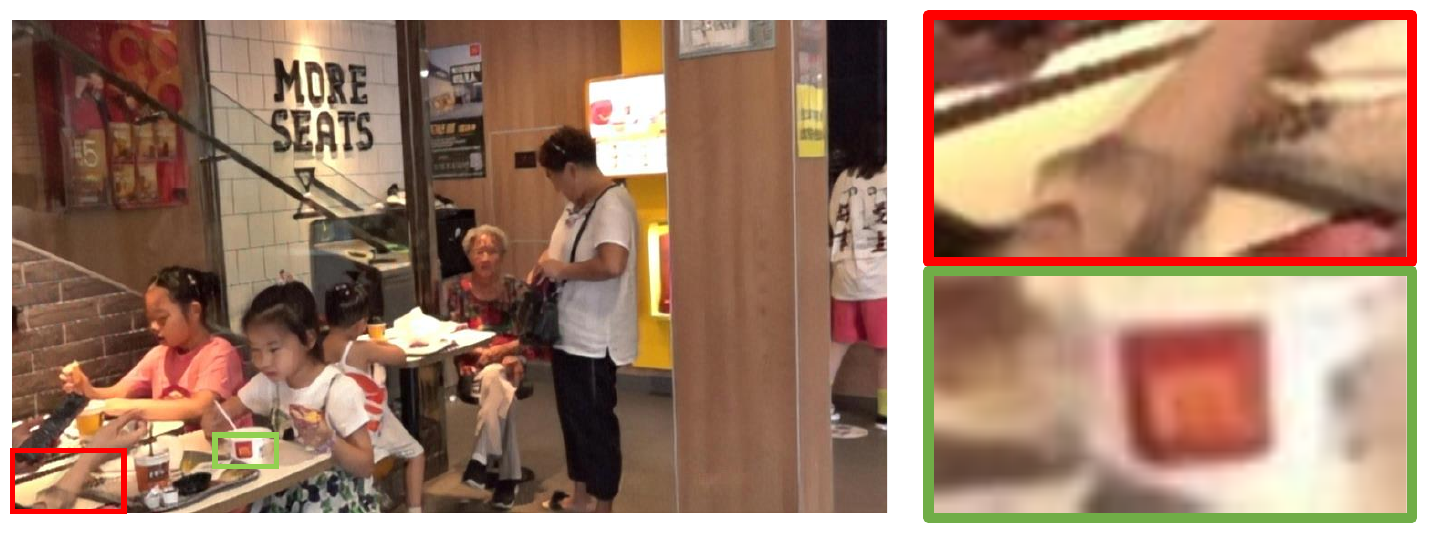}}
   \vspace{-0.15in}
    
    \subfloat[DBGAN \cite{zhang2020deblurring}]{
    \includegraphics[width=0.33\linewidth]{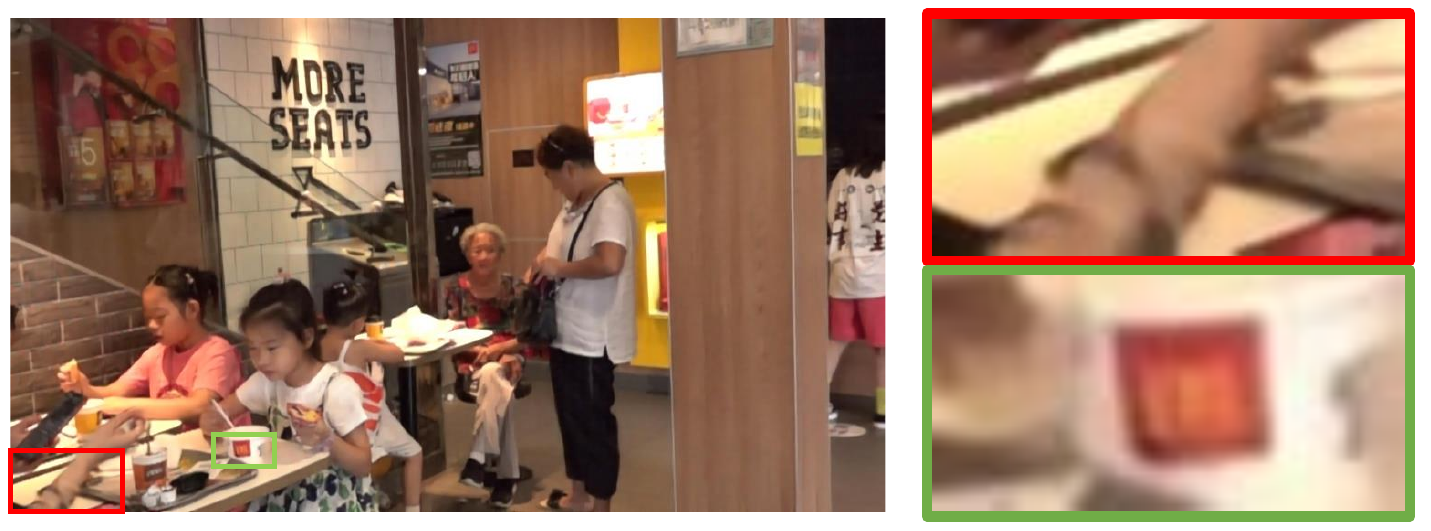}}
    \subfloat[SRN \cite{tao2018scale}]{
    \includegraphics[width=0.33\linewidth]{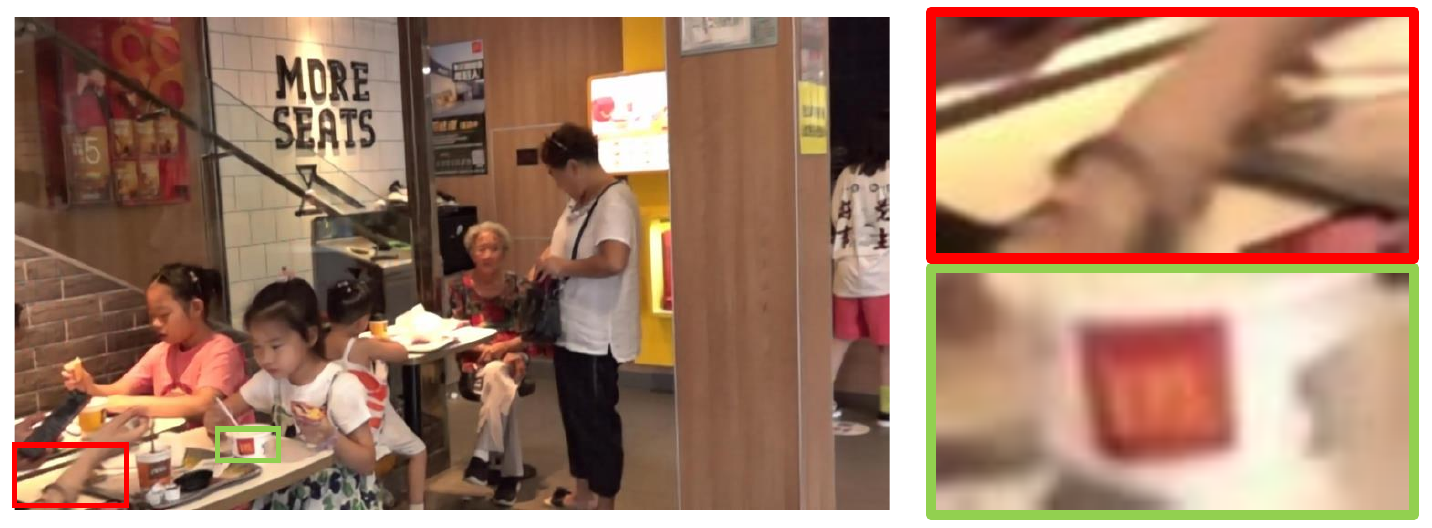}}
    \subfloat[DMPHN \cite{Zhang_2019_CVPR}]{
    \includegraphics[width=0.33\linewidth]{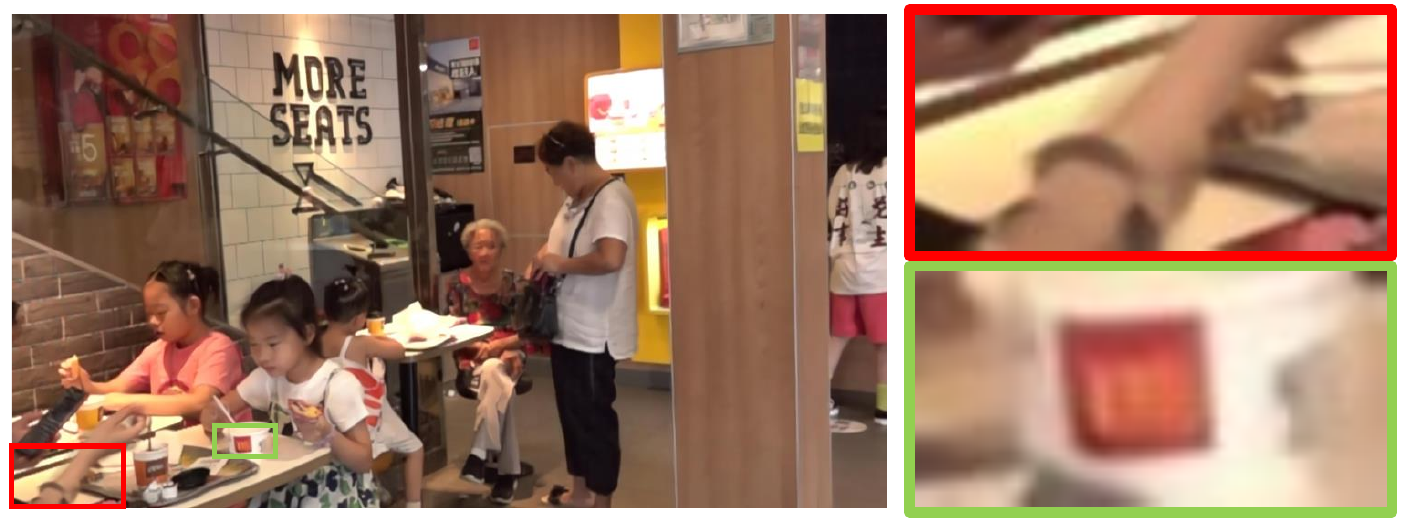}}
\vspace{-0.15in}
    
    \subfloat[MPRNet \cite{zamir2021multi}]{
    \includegraphics[width=0.33\linewidth]{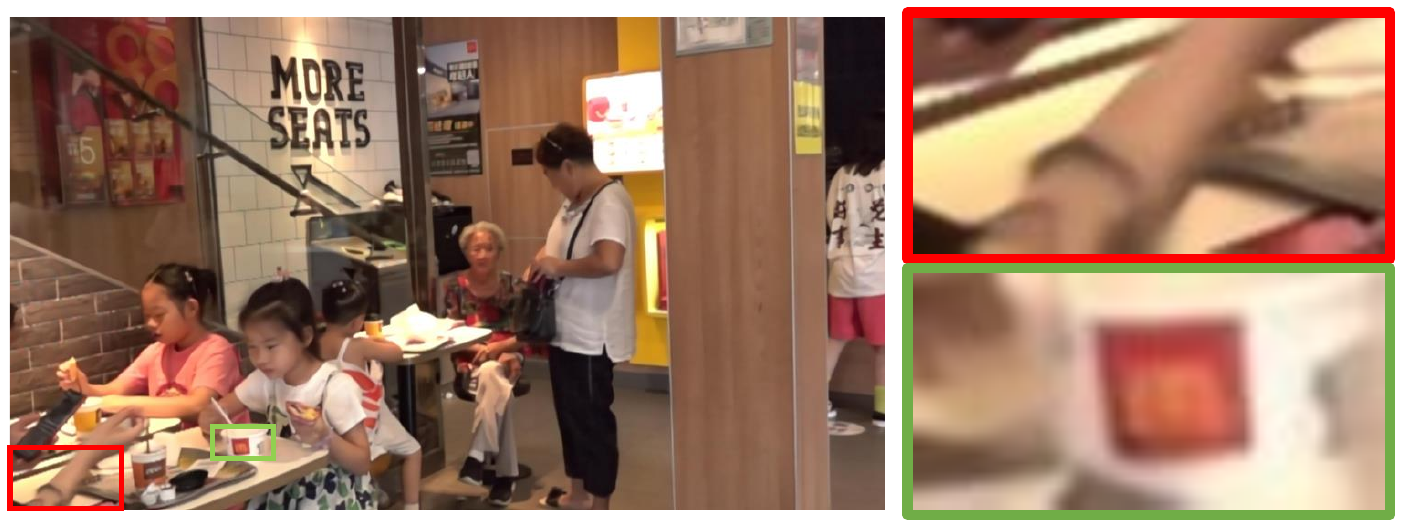}}
    \subfloat[Restormer \cite{zamir2021restormer}]{
    \includegraphics[width=0.33\linewidth]{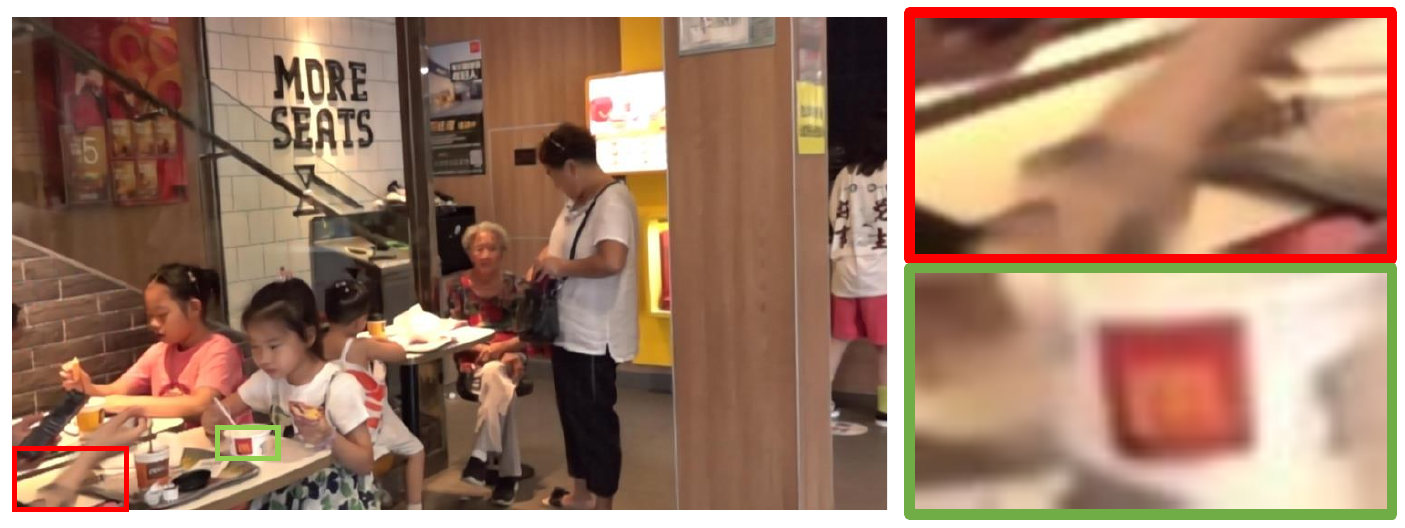}}
    \subfloat[MIMO-UNet \cite{cho2021rethinking}]{
    \includegraphics[width=0.33\linewidth]{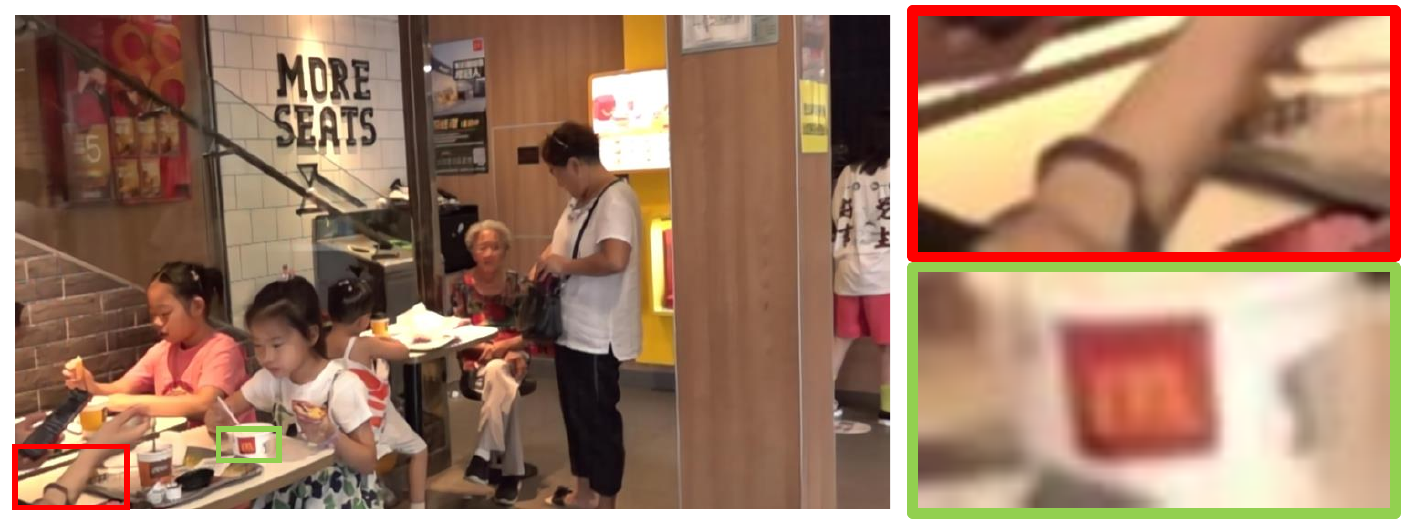}}
    
  \caption{Visual comparisons on the RHM dataset.}
 \label{fig:avg_1} 
 \vspace{-0.2in}
\end{figure*}

\begin{figure*}[t]
  \centering
  \subfloat[Input]{
    \includegraphics[width= 0.33\linewidth]{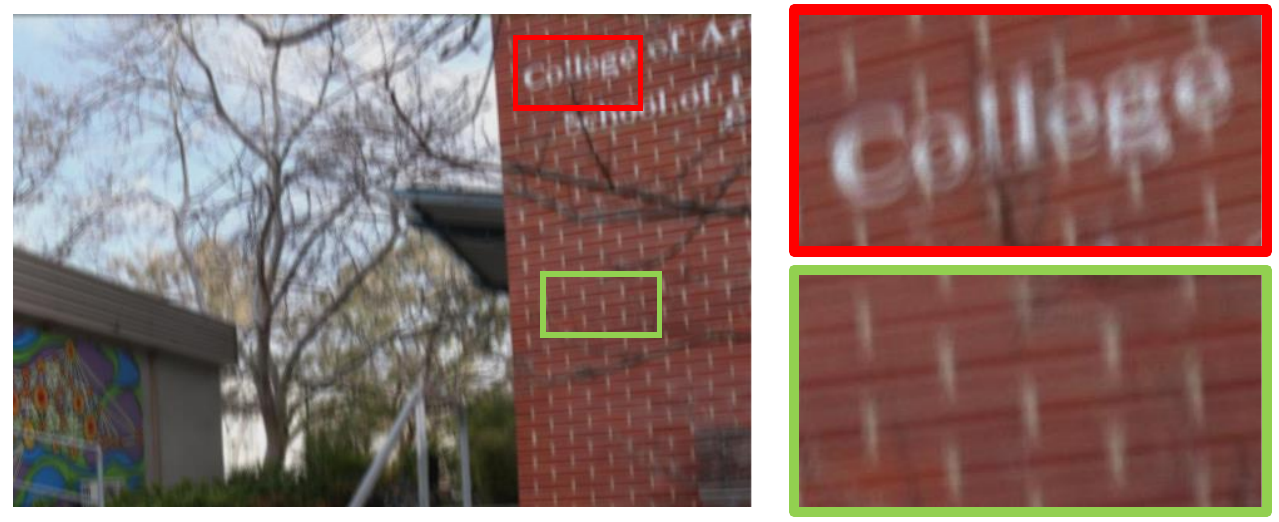}}
  \subfloat[DeepDeblur \cite{nah2017deep}]{
    \includegraphics[width=0.33\linewidth]{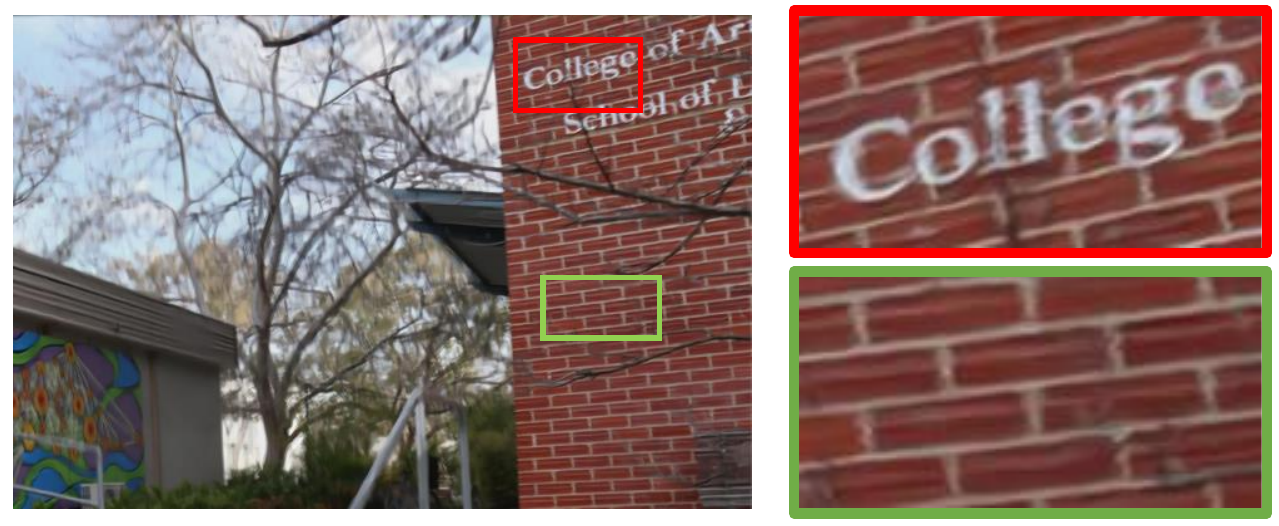}}
    \subfloat[DeblurGAN-v2 \cite{kupyn2019deblurgan}]{
    \includegraphics[width=0.33\linewidth]{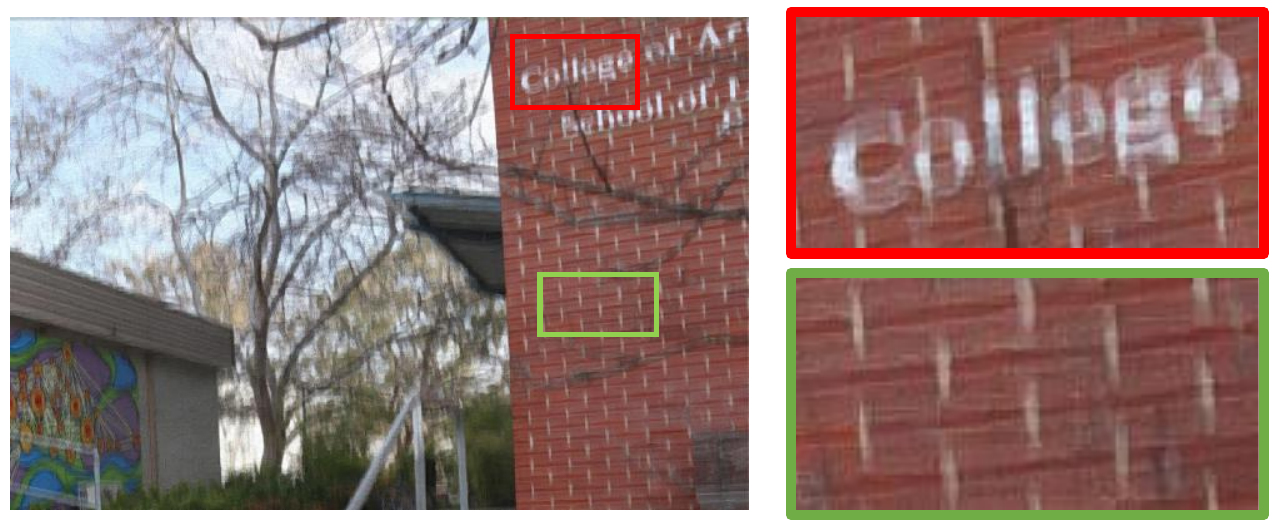}}
    \vspace{-0.15in}

    \subfloat[DBGAN \cite{zhang2020deblurring}]{
    \includegraphics[width=0.33\linewidth]{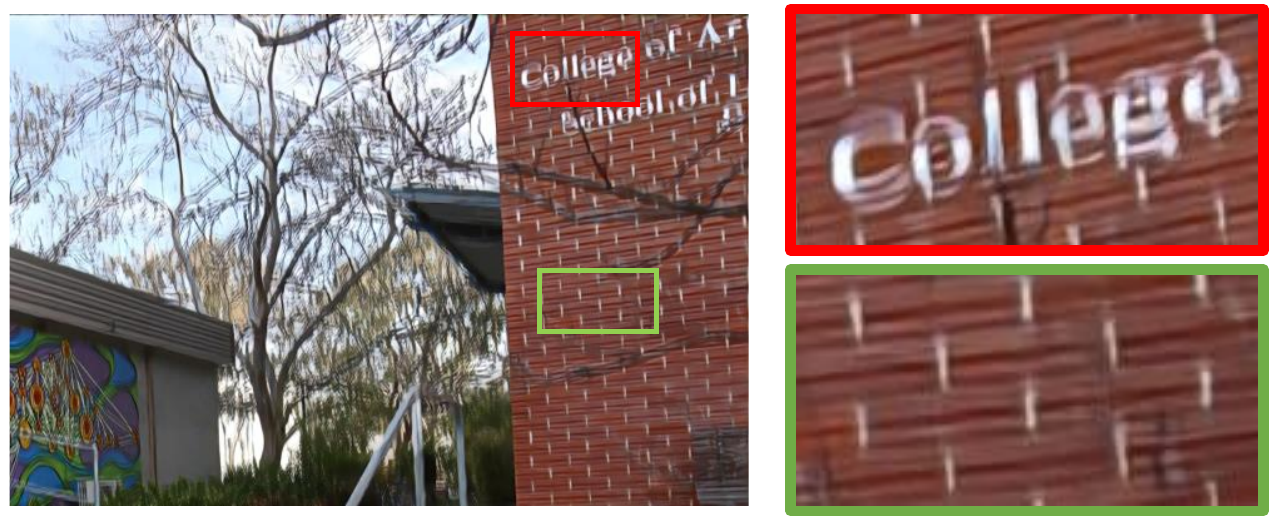}}
    \subfloat[SRN \cite{tao2018scale}]{
    \includegraphics[width=0.33\linewidth]{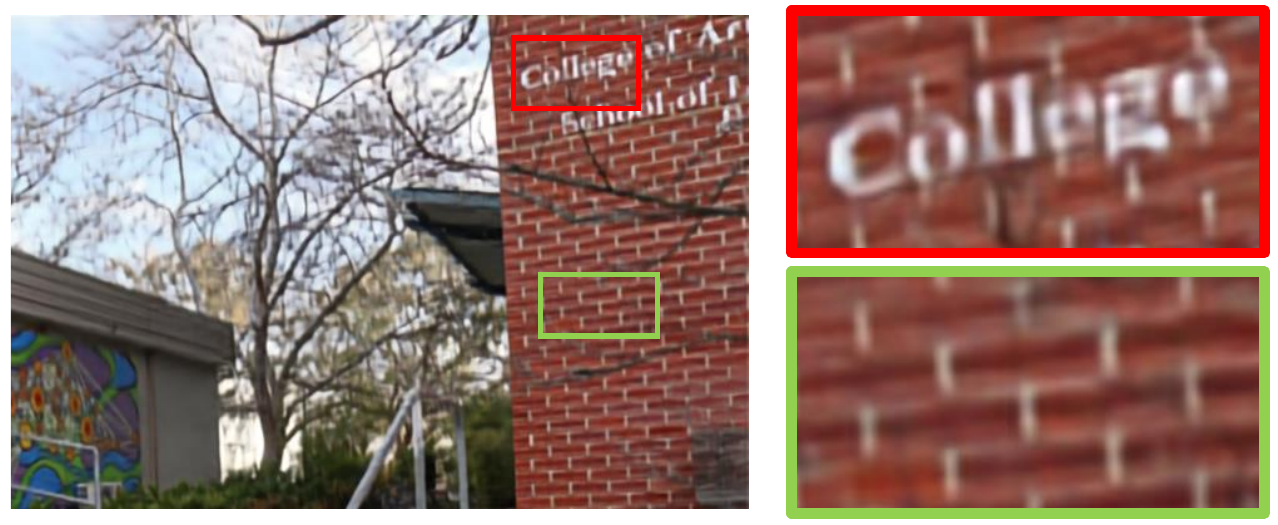}}
    \subfloat[DMPHN \cite{Zhang_2019_CVPR}]{
    \includegraphics[width=0.33\linewidth]{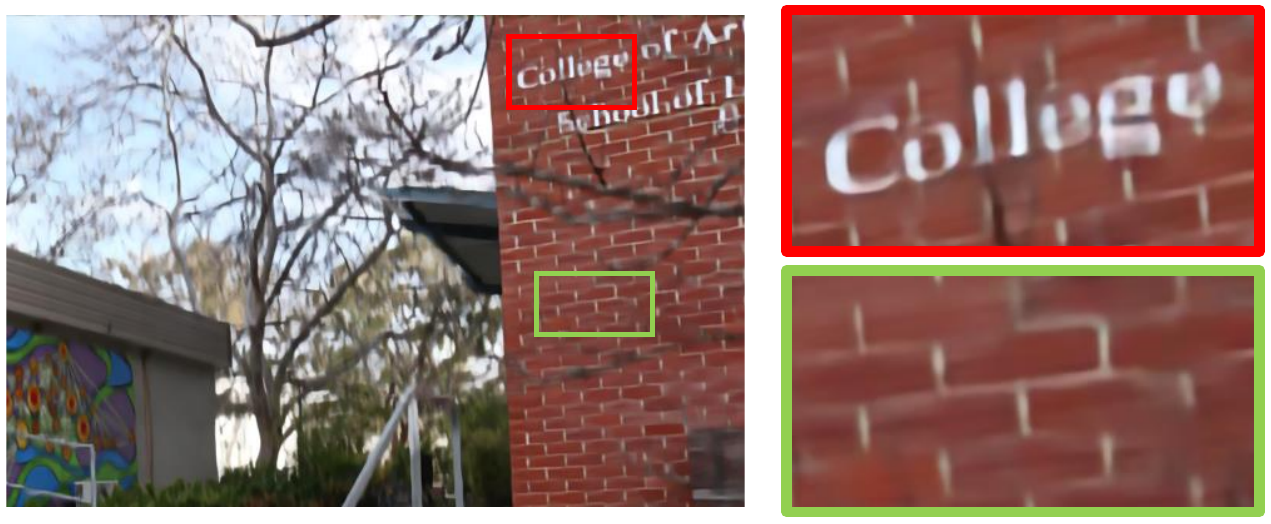}}
    \vspace{-0.15in}

    \subfloat[MPRNet \cite{zamir2021multi}]{
    \includegraphics[width=0.33\linewidth]{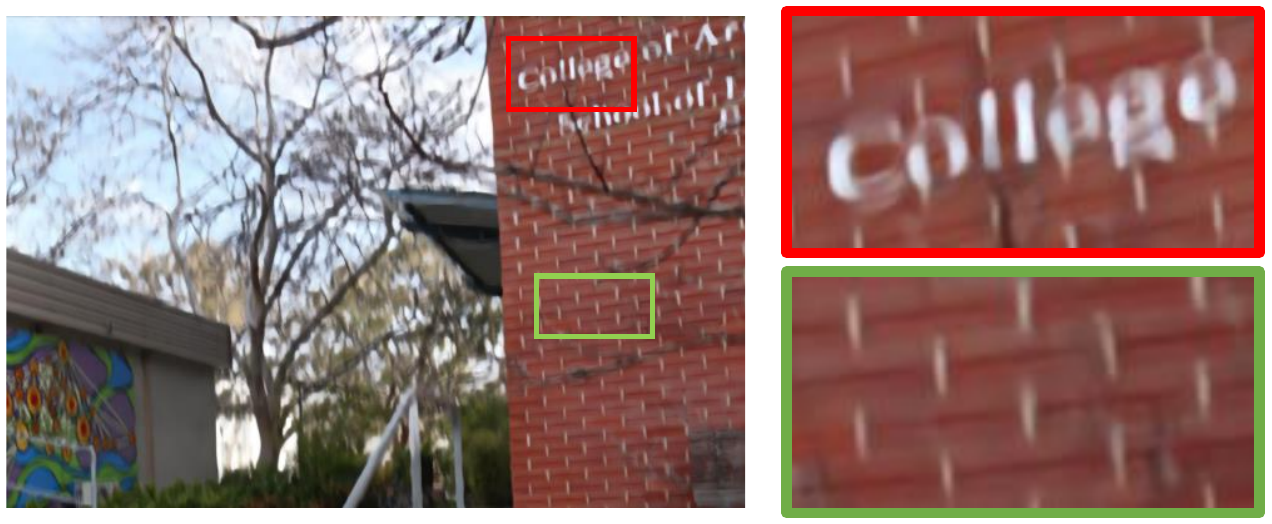}}
    \subfloat[Restormer \cite{zamir2021restormer}]{
    \includegraphics[width=0.33\linewidth]{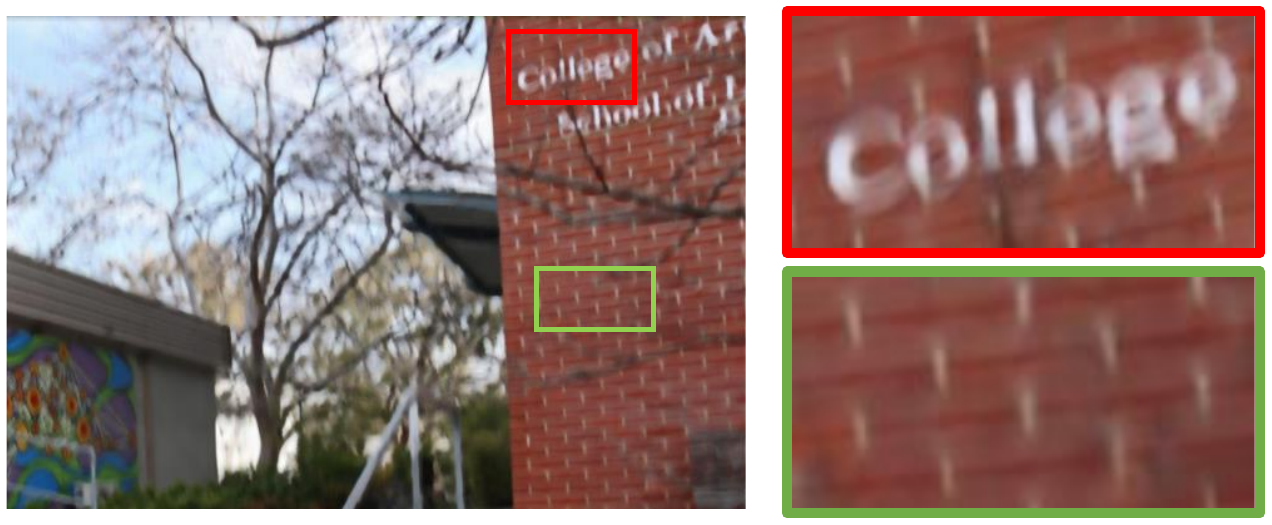}}
    \subfloat[MIMO-UNet \cite{cho2021rethinking}]{
    \includegraphics[width=0.33\linewidth]{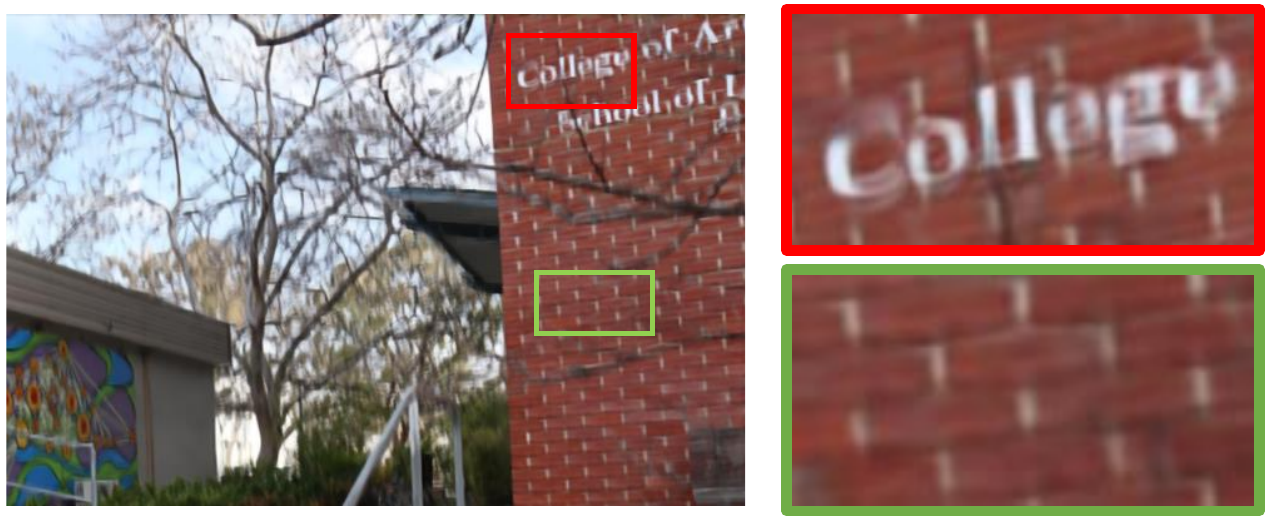}}
    
  \caption{Visual comparisons on the proposed UHDM set.}
 \label{fig:conv_1} 
\end{figure*}

\begin{figure*}[t]
  \centering
  \subfloat[Input]{
    \includegraphics[width= 0.33\linewidth]{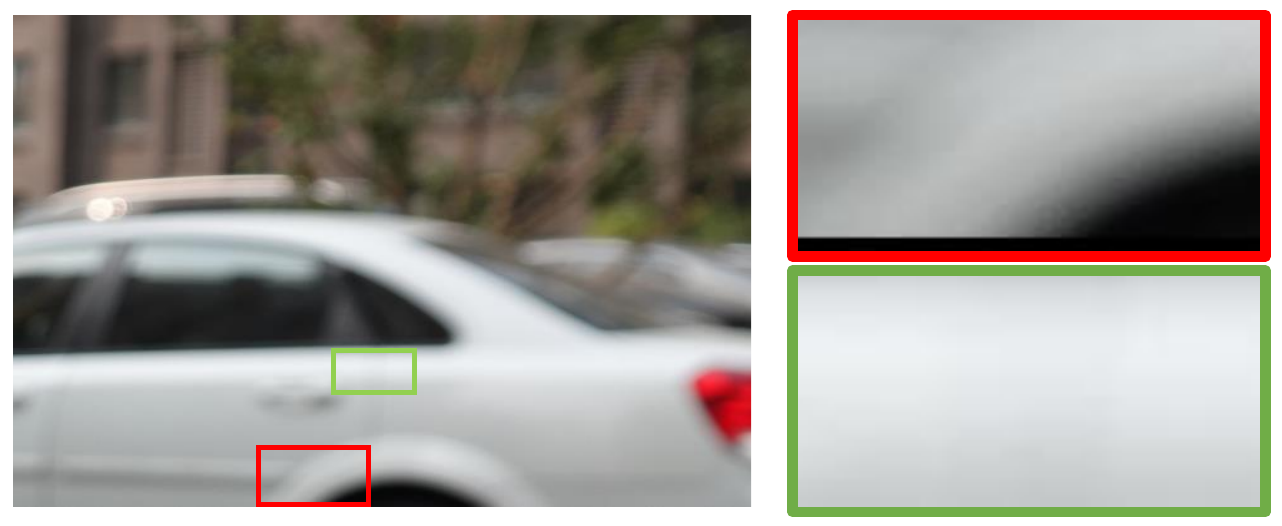}}
  \subfloat[DeepDeblur \cite{nah2017deep}]{
    \includegraphics[width=0.33\linewidth]{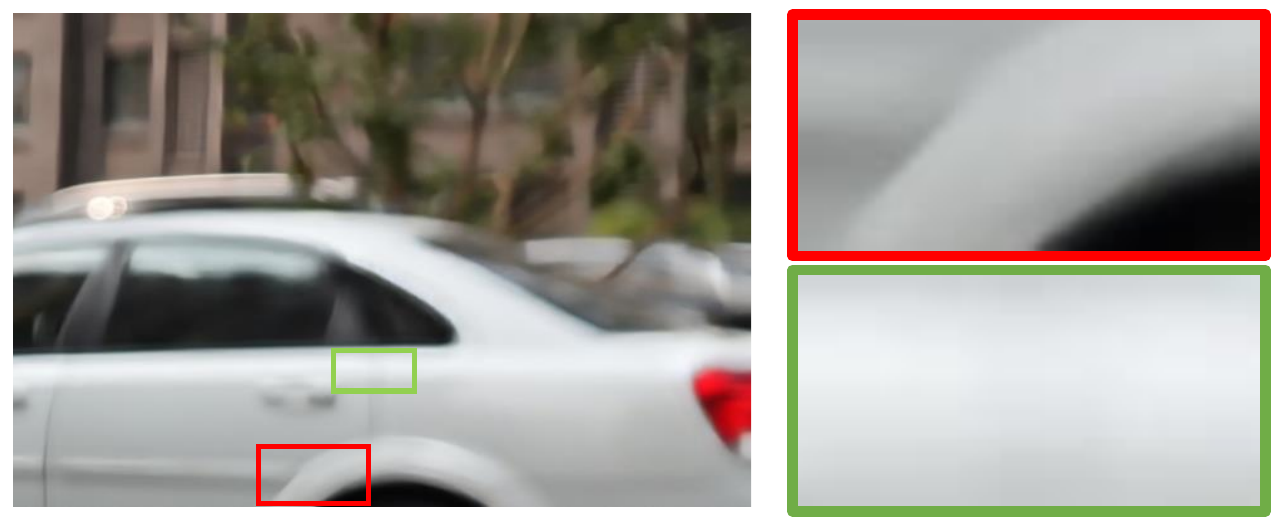}}
    \subfloat[DeblurGAN-v2 \cite{kupyn2019deblurgan}]{
    \includegraphics[width=0.33\linewidth]{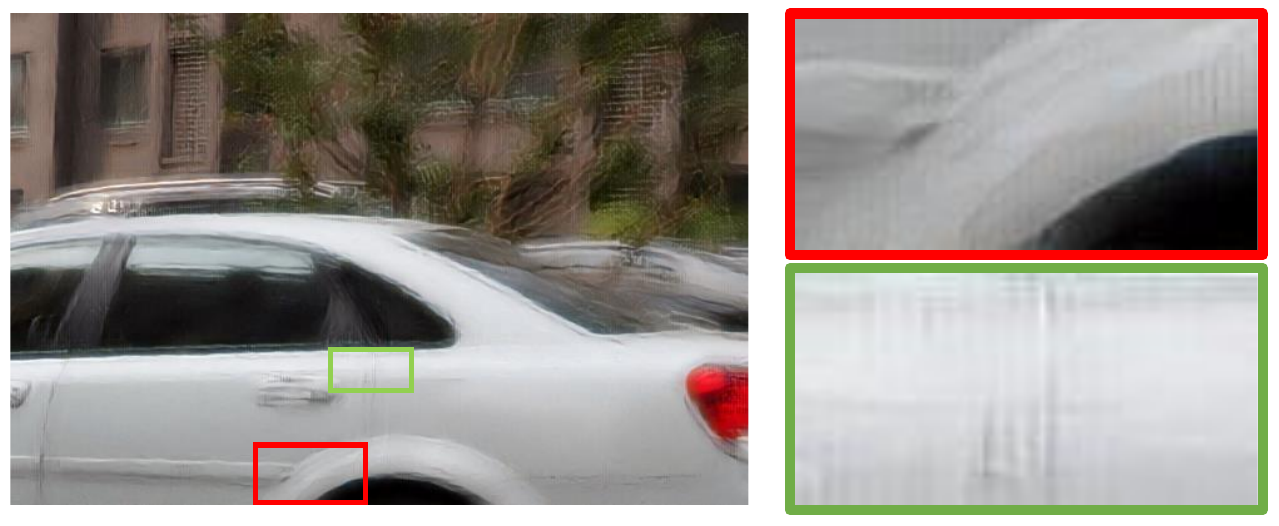}}
\vspace{-0.15in}

    \subfloat[DBGAN \cite{zhang2020deblurring}]{
    \includegraphics[width=0.33\linewidth]{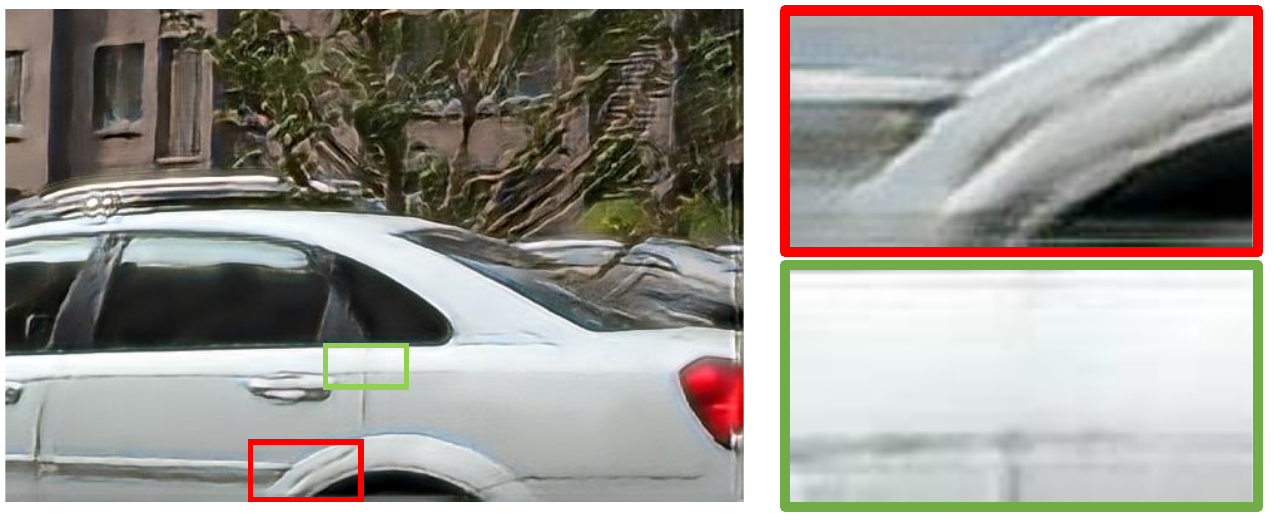}}
    \subfloat[SRN \cite{tao2018scale}]{
    \includegraphics[width=0.33\linewidth]{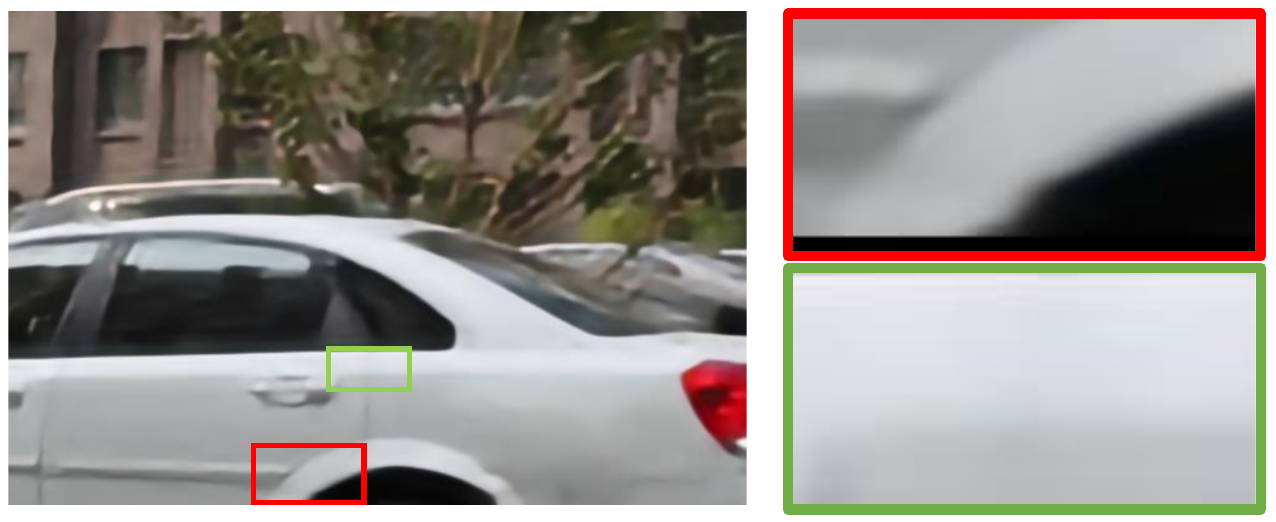}}
    \subfloat[DMPHN \cite{Zhang_2019_CVPR}]{
    \includegraphics[width=0.33\linewidth]{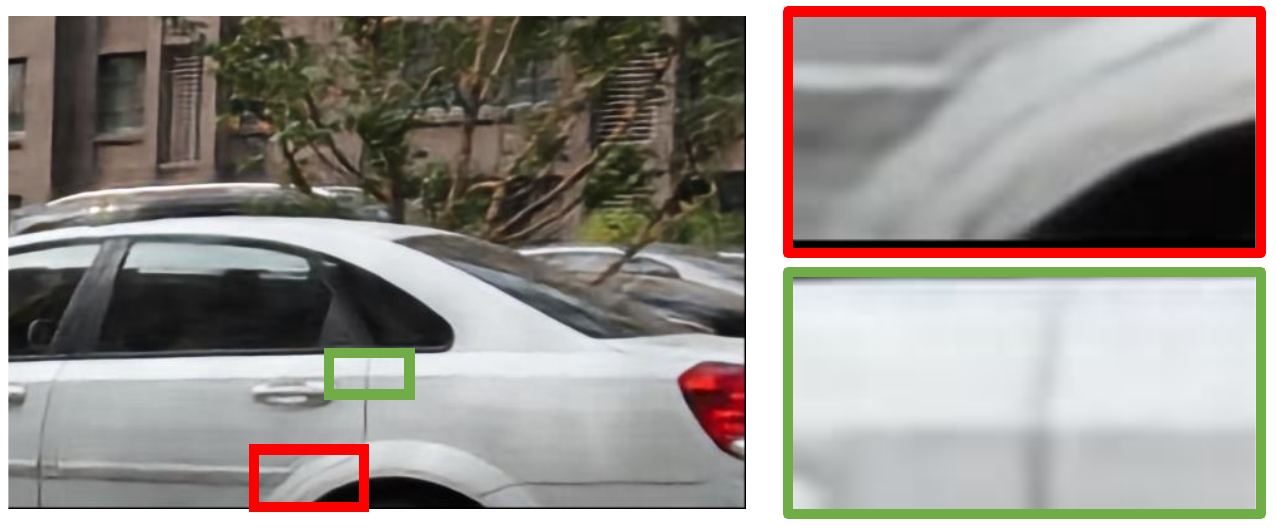}}
\vspace{-0.15in}

    \subfloat[MPRNet \cite{zamir2021multi}]{
    \includegraphics[width=0.33\linewidth]{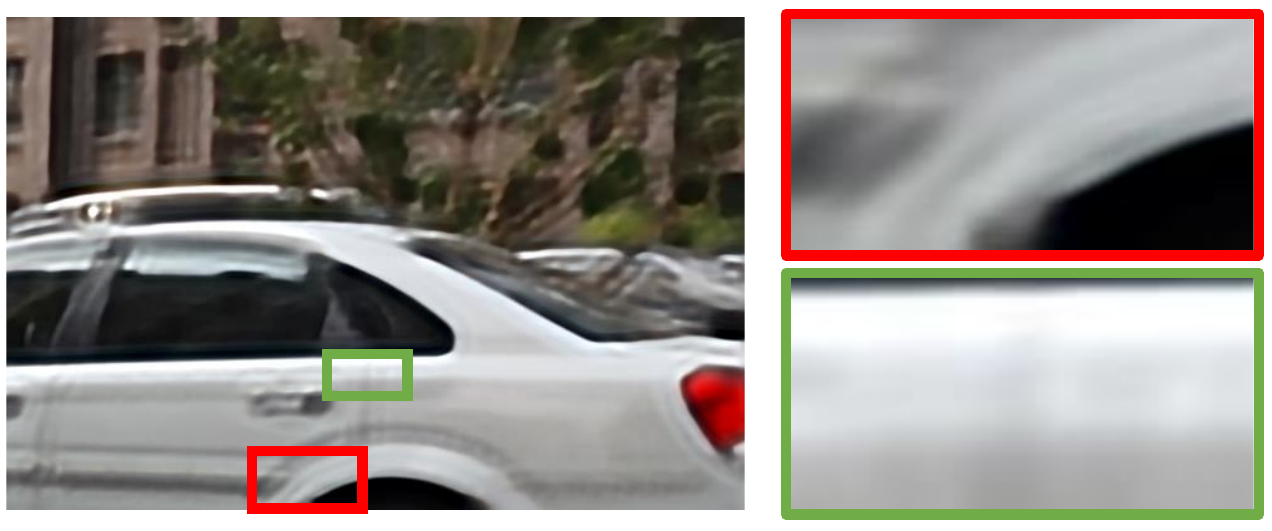}}
    \subfloat[Restormer \cite{zamir2021restormer}]{
    \includegraphics[width=0.33\linewidth]{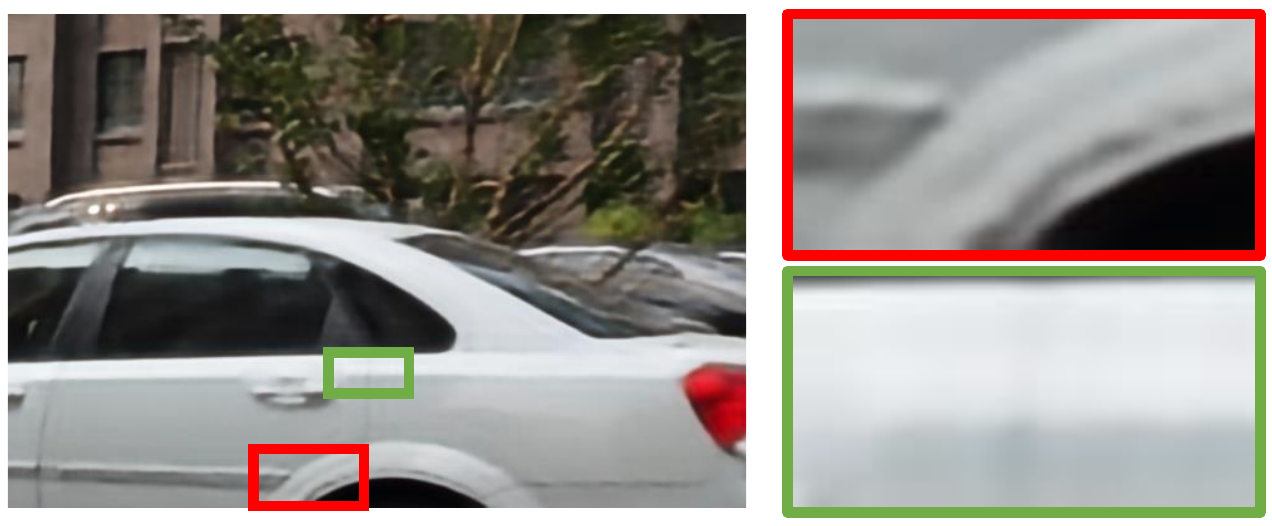}}
    \subfloat[MIMO-UNet \cite{cho2021rethinking}]{
    \includegraphics[width=0.33\linewidth]{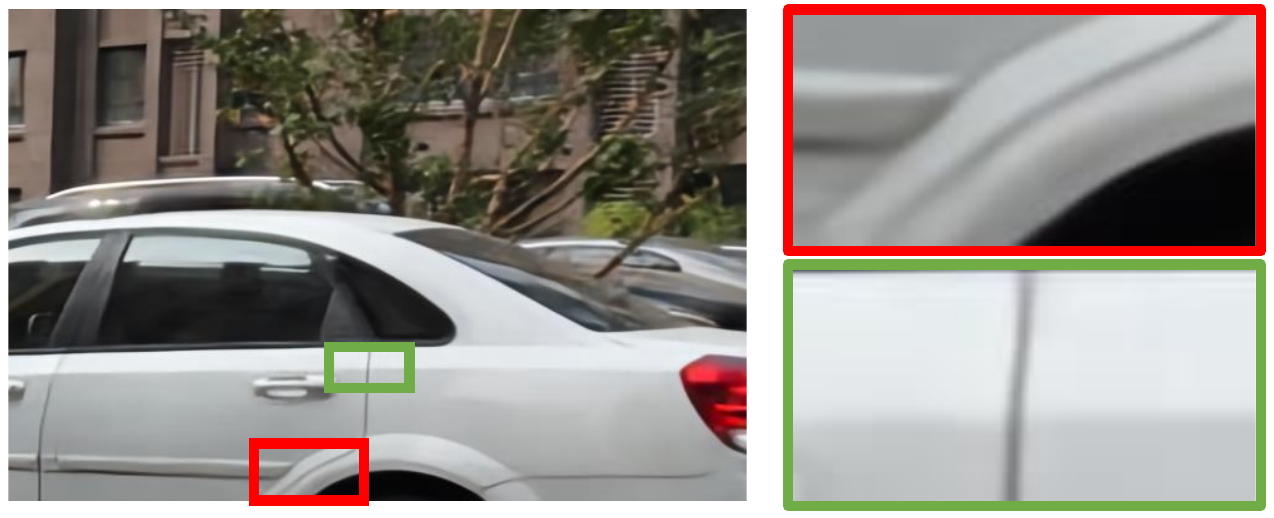}}
  \caption{Visual comparisons on the proposed LSD set.}
 \label{fig:defocus_1} 
\end{figure*}

\begin{figure*}[h]
  \centering
  \subfloat[Input]{
    \includegraphics[width= 0.33\linewidth]{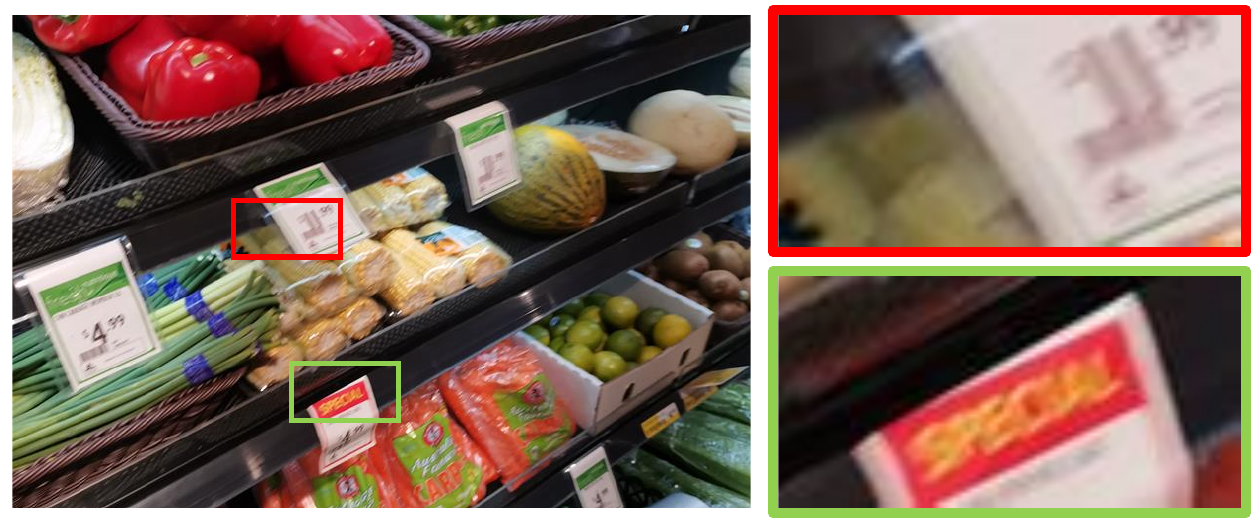}}
  \subfloat[DeepDeblur \cite{nah2017deep}]{
    \includegraphics[width=0.33\linewidth]{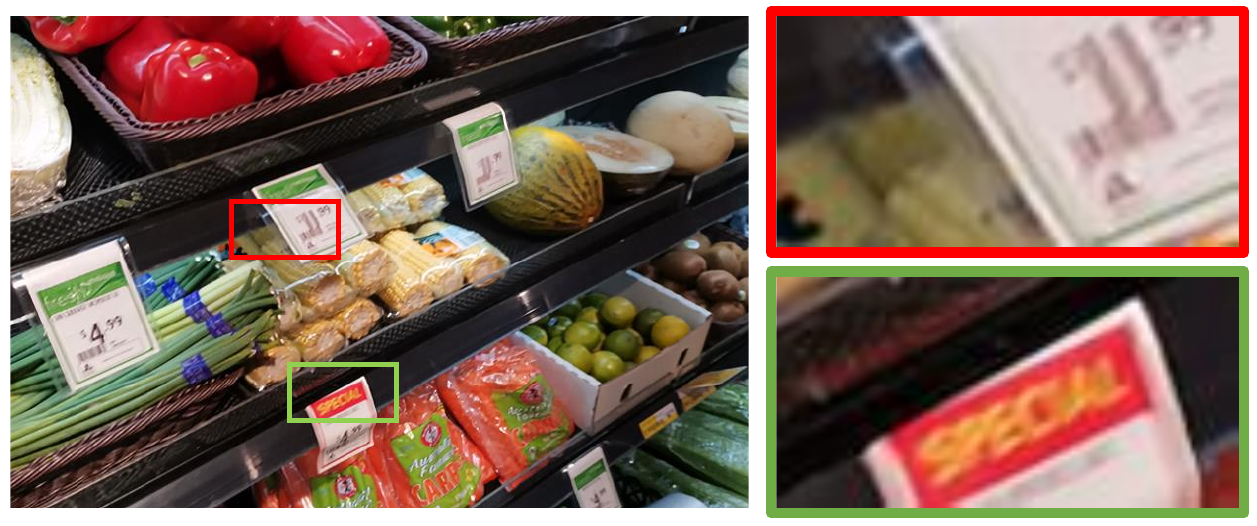}}
    \subfloat[DeblurGAN-v2 \cite{kupyn2019deblurgan}]{
    \includegraphics[width=0.33\linewidth]{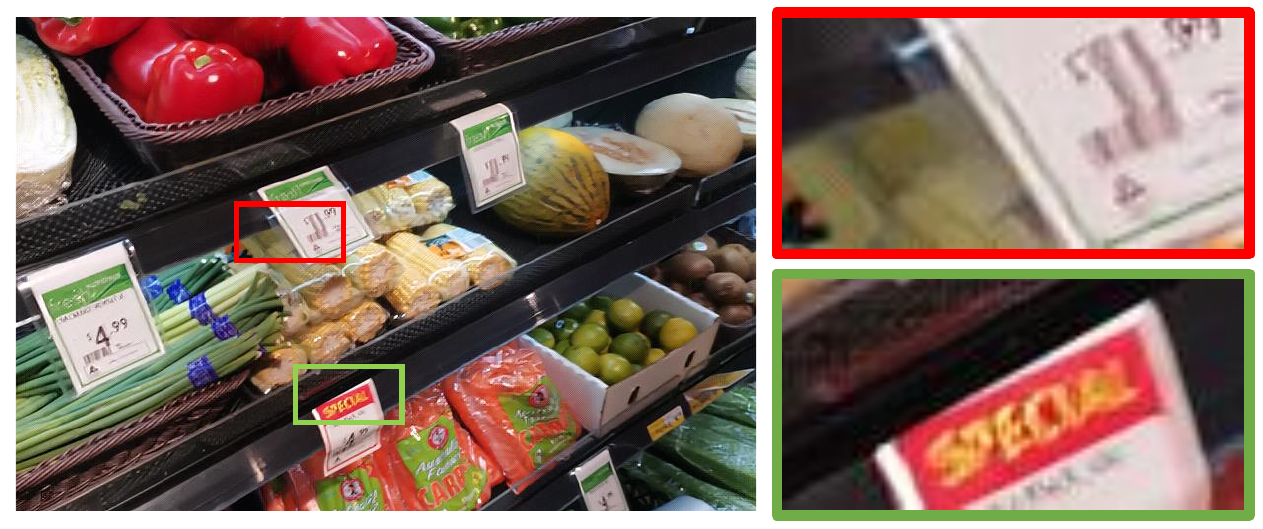}}
\vspace{-0.15in}

    \subfloat[DBGAN \cite{zhang2020deblurring}]{
    \includegraphics[width=0.33\linewidth]{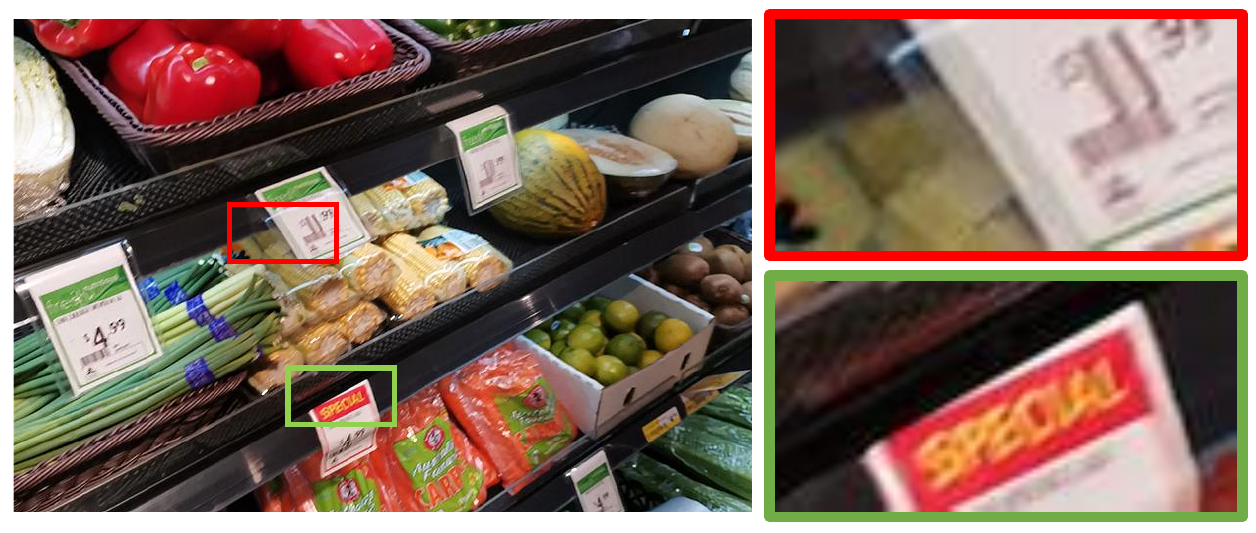}}
    \subfloat[SRN \cite{tao2018scale}]{
    \includegraphics[width=0.33\linewidth]{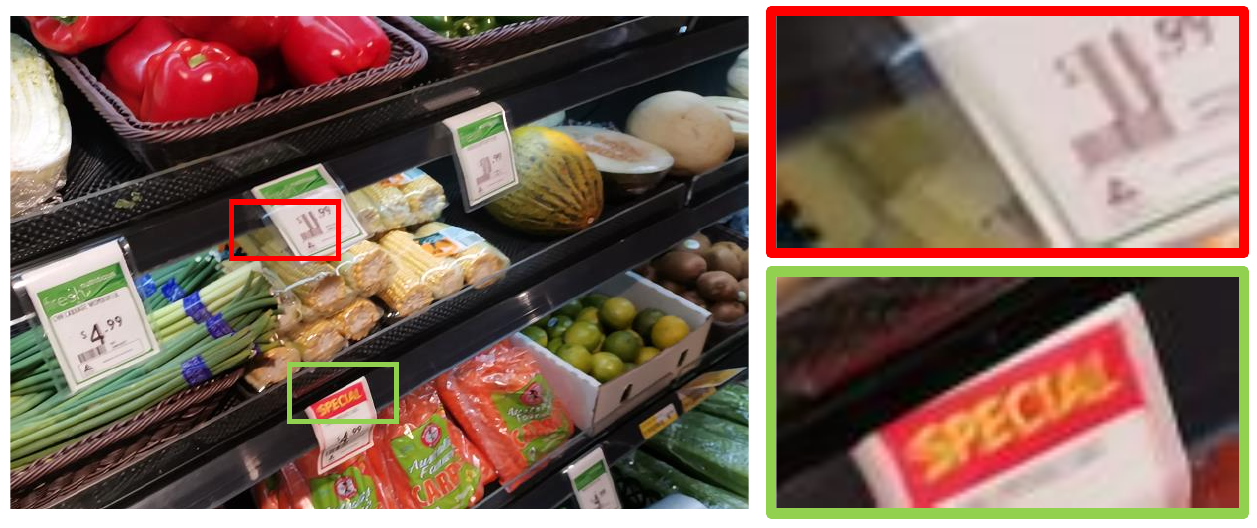}}
    \subfloat[DMPHN \cite{Zhang_2019_CVPR}]{
    \includegraphics[width=0.33\linewidth]{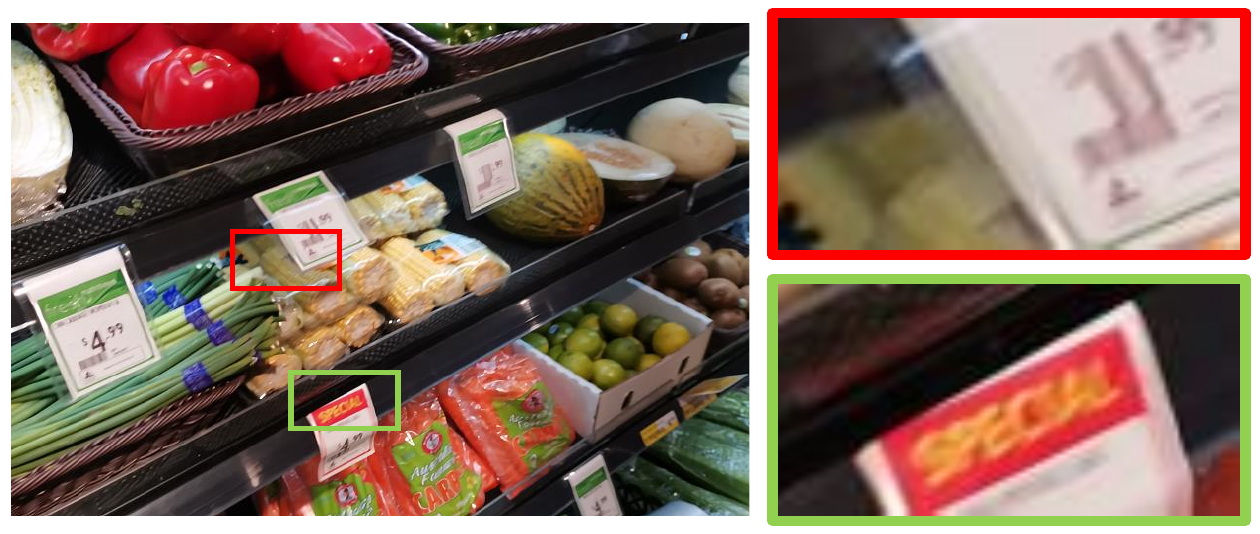}}
\vspace{-0.15in}

    \subfloat[MPRNet \cite{zamir2021multi}]{
    \includegraphics[width=0.33\linewidth]{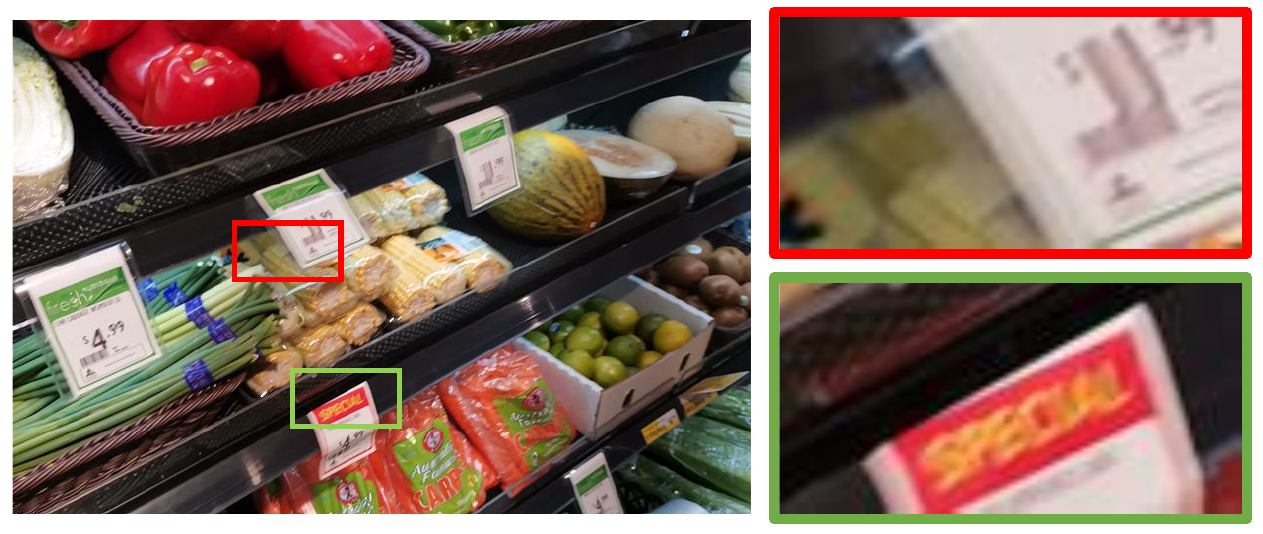}}
    \subfloat[Restormer \cite{zamir2021restormer}]{
    \includegraphics[width=0.33\linewidth]{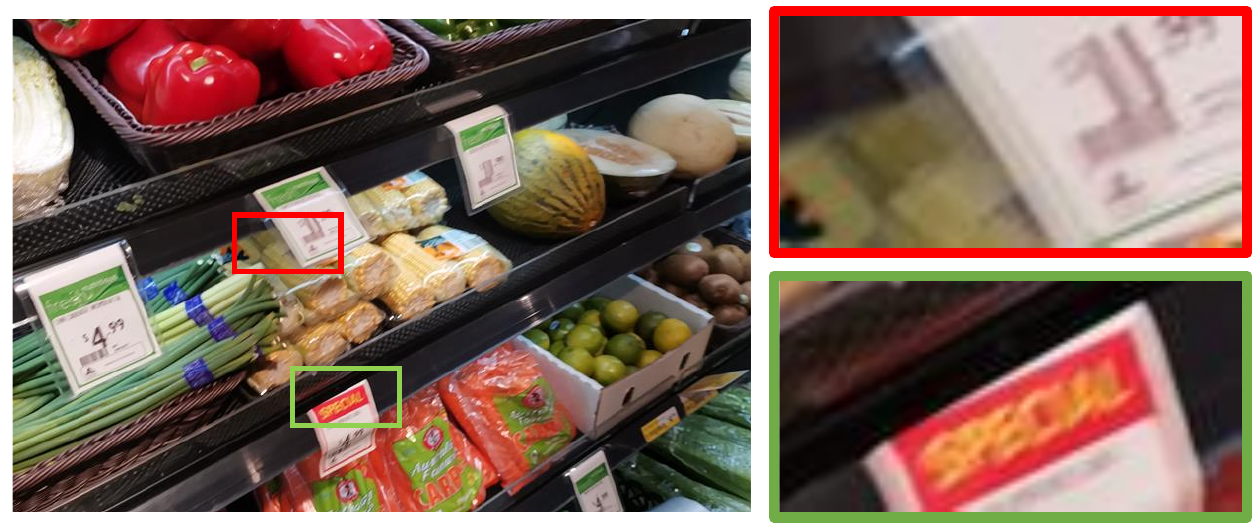}}
    \subfloat[MIMO-UNet \cite{cho2021rethinking}]{
    \includegraphics[width=0.33\linewidth]{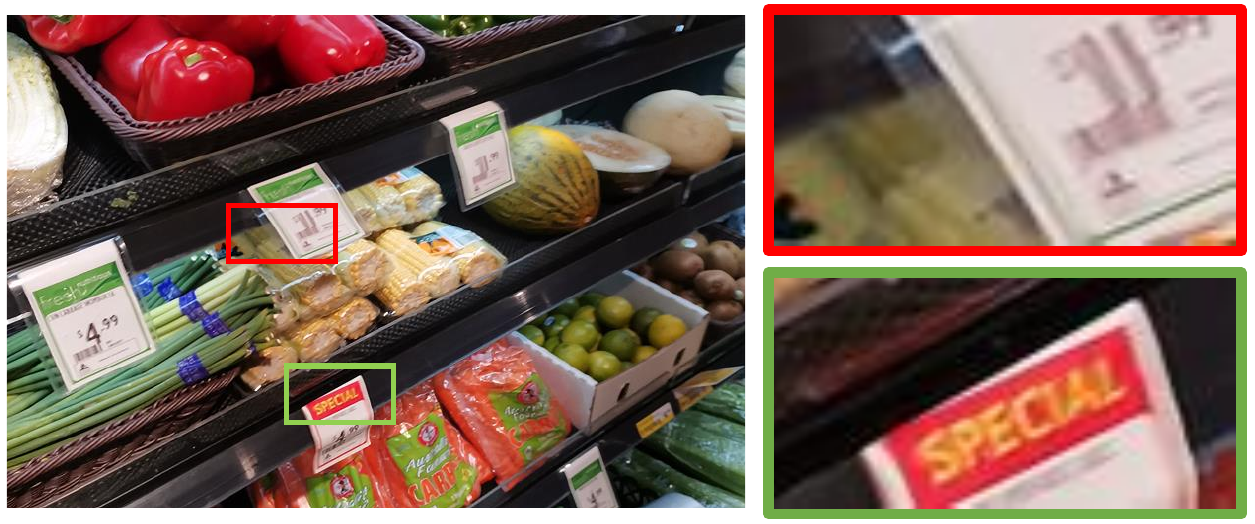}}
  \caption{Visual comparisons on the proposed RMBQ set.}
 \label{fig:real_1} 
\end{figure*}

\begin{figure*}[h]
  \centering
  \subfloat[Input]{
    \includegraphics[width= 0.33\linewidth]{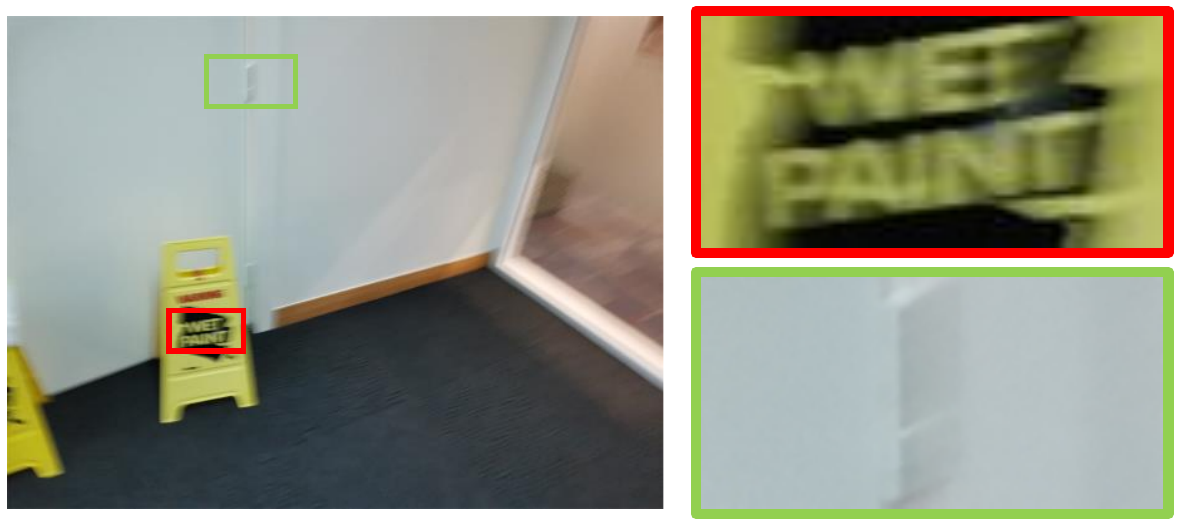}}
  \subfloat[DeepDeblur \cite{nah2017deep}]{
    \includegraphics[width=0.33\linewidth]{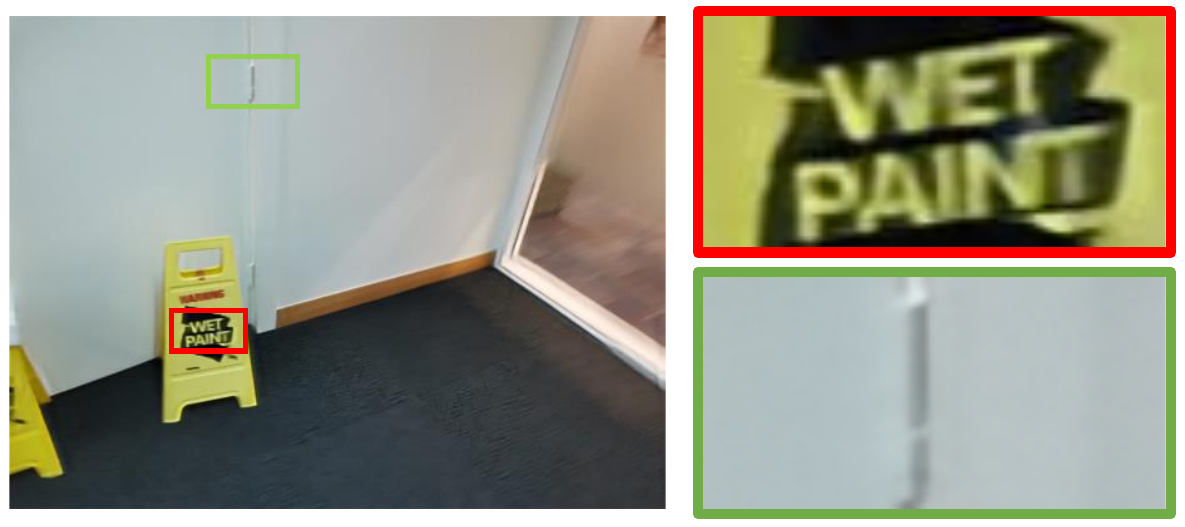}}
    \subfloat[DeblurGAN-v2 \cite{kupyn2019deblurgan}]{
    \includegraphics[width=0.33\linewidth]{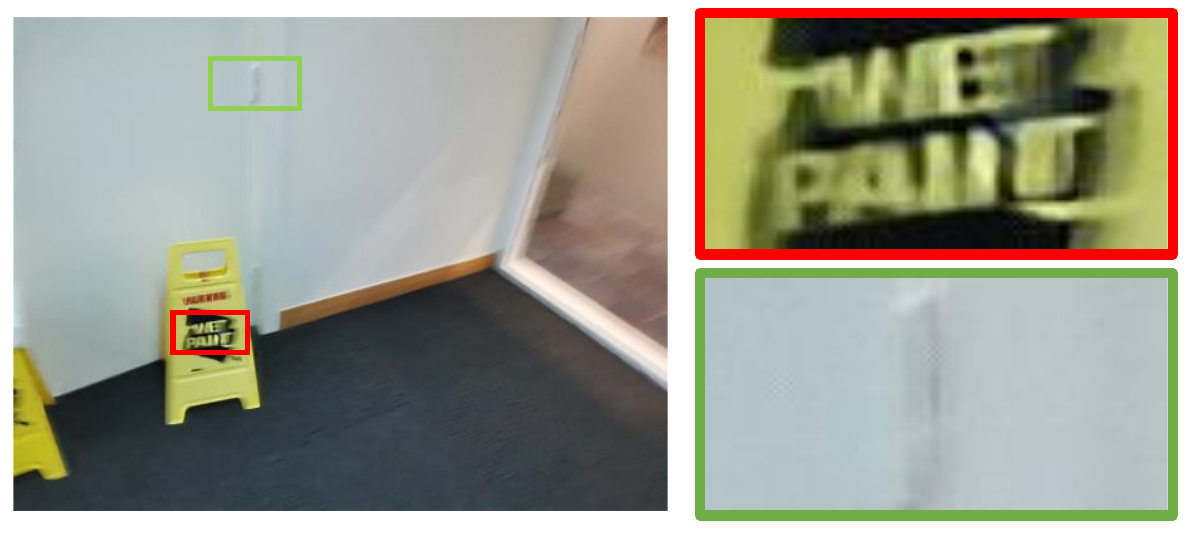}}
\vspace{-0.15in}

    \subfloat[DBGAN \cite{zhang2020deblurring}]{
    \includegraphics[width=0.33\linewidth]{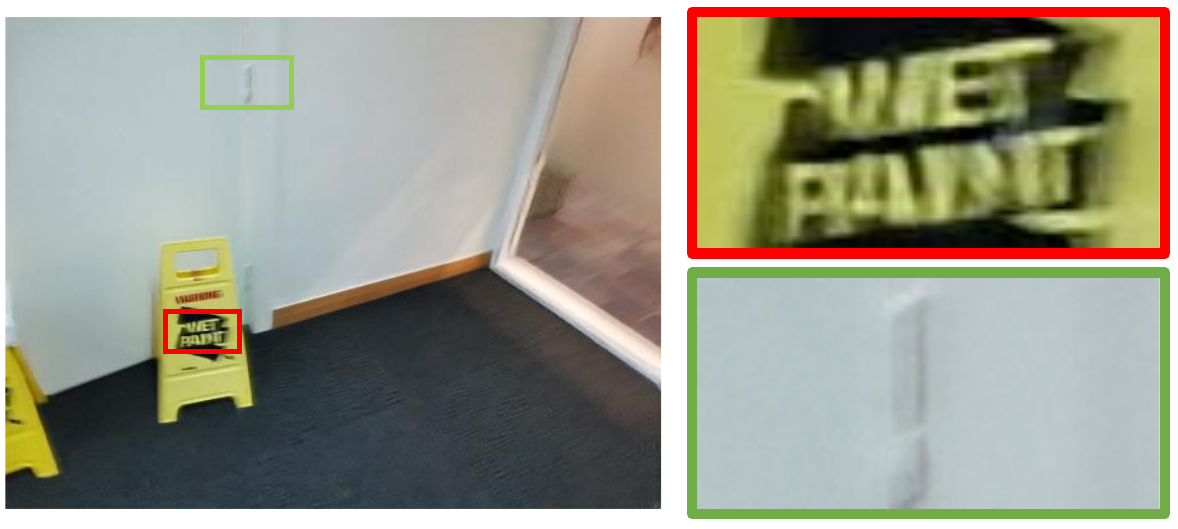}}
    \subfloat[SRN \cite{tao2018scale}]{
    \includegraphics[width=0.33\linewidth]{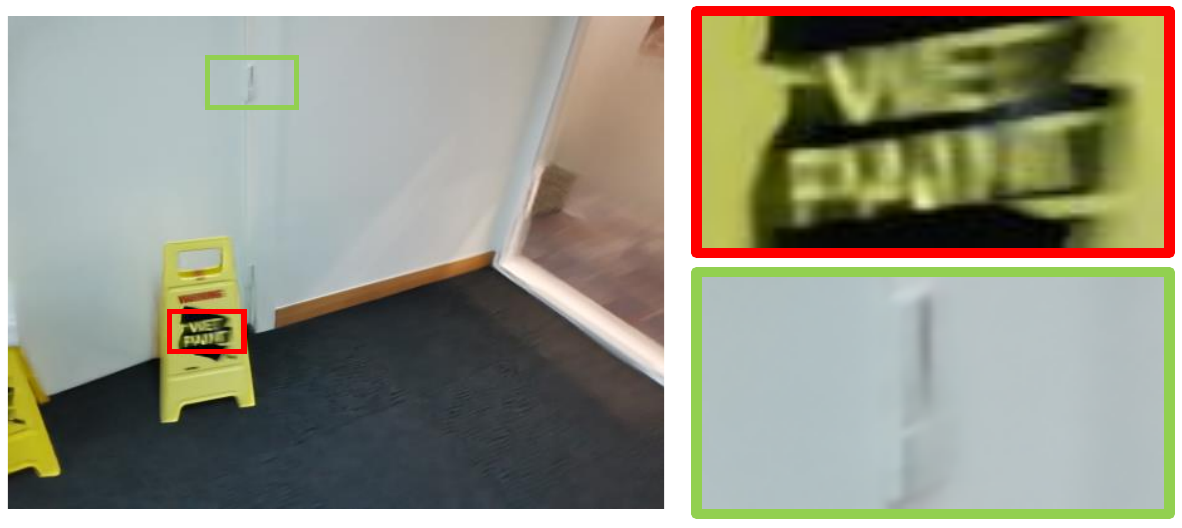}}
    \subfloat[DMPHN \cite{Zhang_2019_CVPR}]{
    \includegraphics[width=0.33\linewidth]{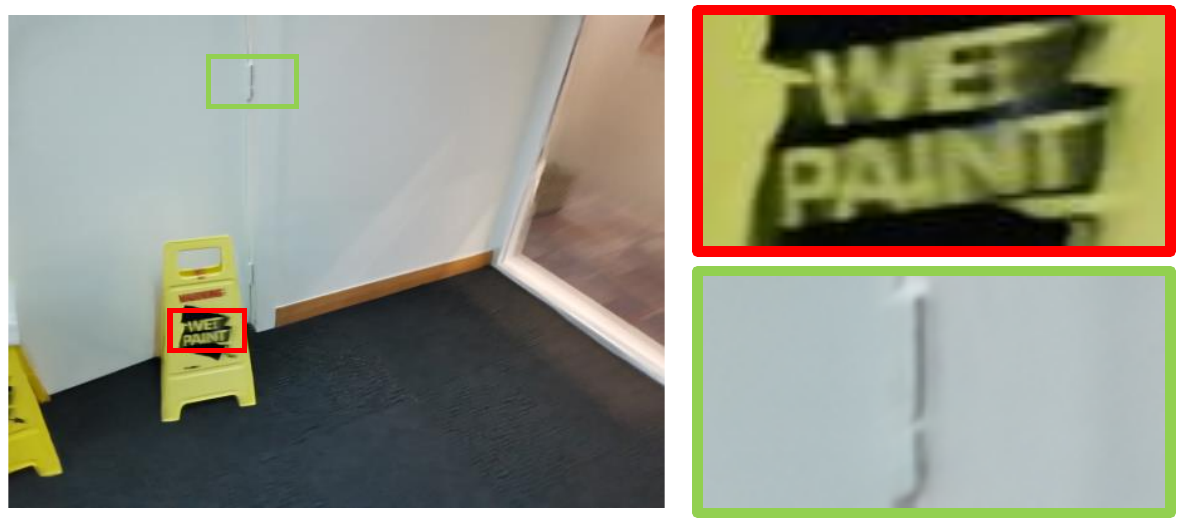}}
\vspace{-0.15in}

    \subfloat[MPRNet \cite{zamir2021multi}]{
    \includegraphics[width=0.33\linewidth]{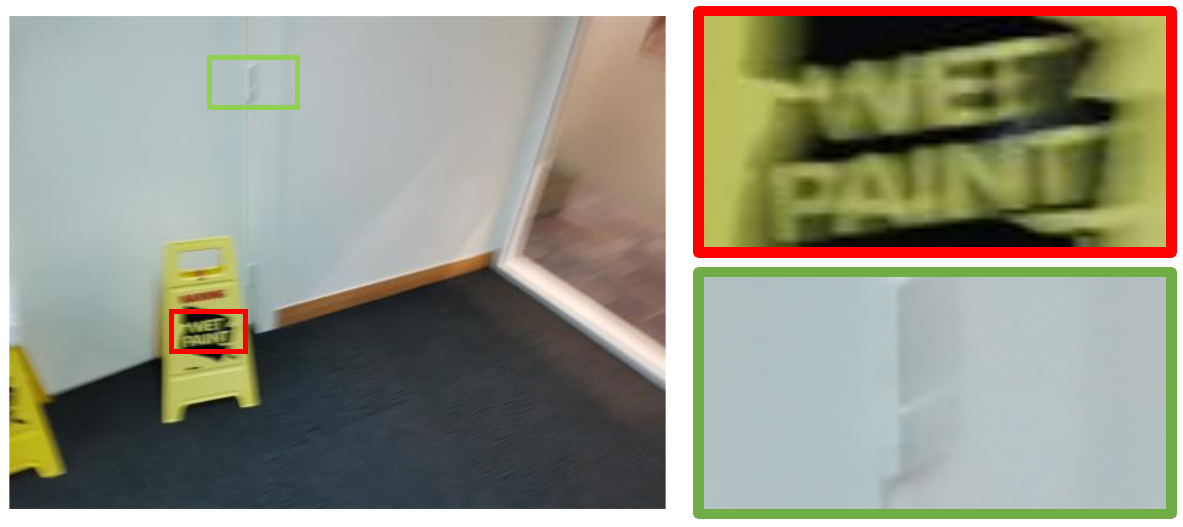}}
    \subfloat[Restormer \cite{zamir2021restormer}]{
    \includegraphics[width=0.33\linewidth]{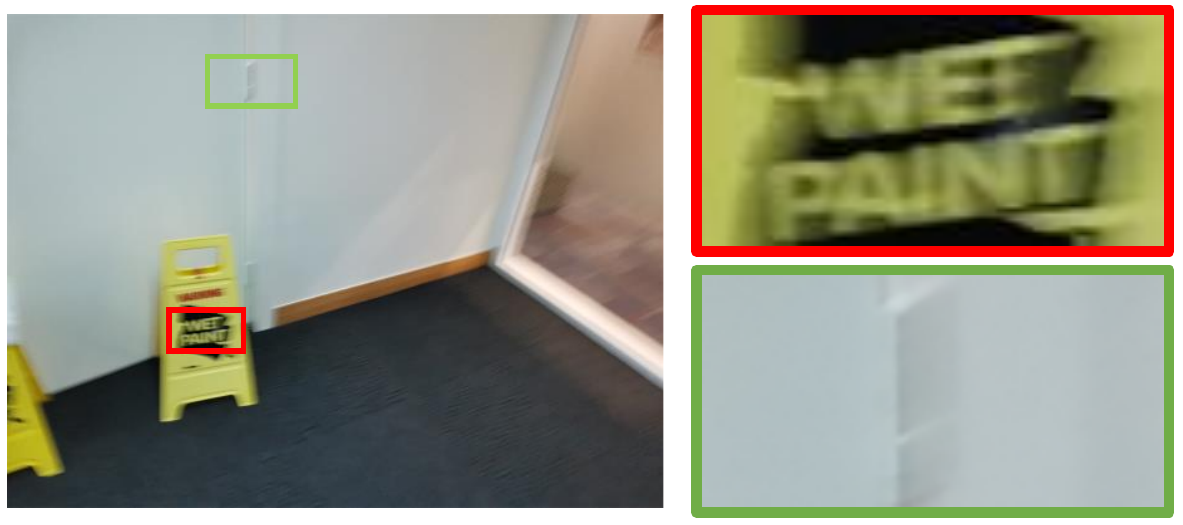}}
    \subfloat[MIMO-UNet \cite{cho2021rethinking}]{
    \includegraphics[width=0.33\linewidth]{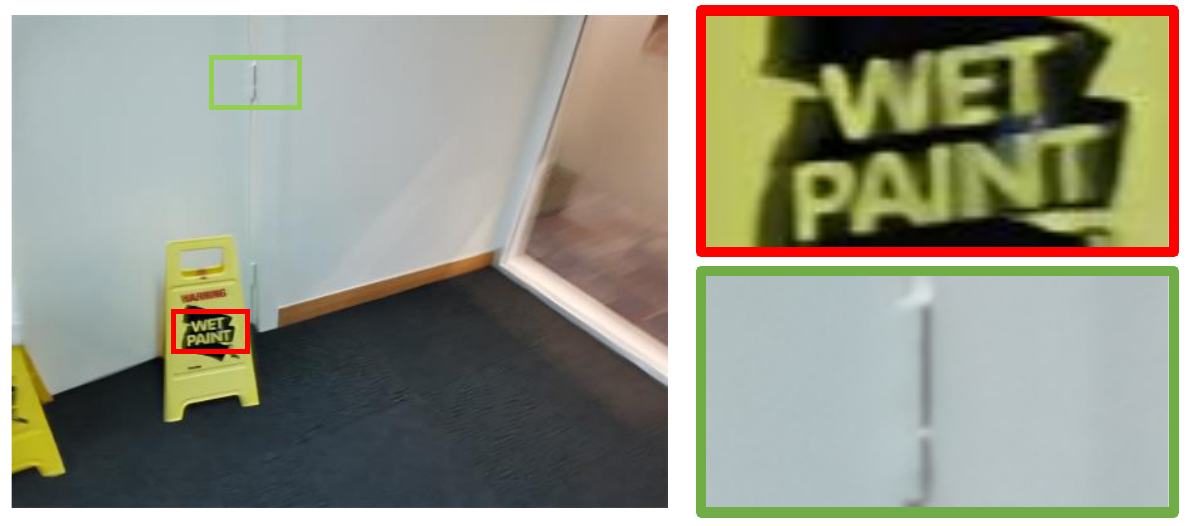}}
  \caption{Visual comparisons on the proposed RMBQ set.}
 \label{fig:real_2} 
\end{figure*}

\noindent \textbf{RMBQ Subset.}
The above three sets contain different kinds of real and synthesized blurry images.
In the real world, blur artifacts can be more complex and difficult to approximate.  
For instance, real-world blur is caused by multiple reasons, such as the blur caused by both camera shake and object movement.
Thus, it is difficult to guarantee the generalization of models trained with images containing only a specific kind of blur. 
Therefore, we capture another set of blurry images with various devices, including high-end digital cameras and mobile phones (iPhone, Samsung, and Huawei). 
There are $10,000$ images in this real mixed blurry qualitative set, RMBQ. 
This set is designed only for qualitative evaluation, as no sharp ground truth images are available.

\section{Benchmarking and Analysis}
\label{Experiments}

In this section, we present the benchmarking results for established deblurring approaches using the proposed MC-Blur dataset. 
We begin by introducing the evaluated deblurring methods in Sec. \ref{experiment: methods} and assess their performance across various categories of blurry images in Sec. \ref{experiment: mc}. 
Subsequently, we conduct an efficiency analysis on UHD blurry images in Sec. \ref{exp:efficiency}. 
Following that, we present benchmark studies involving cross-dataset learning in Sec. \ref{exp:dataset_comparison}, \ref{exp:dataset_comparison_defocus}, and \ref{exp:dataset_comparison_different_blur}. 
Finally, we summarize the key insights derived from these benchmarking experiments in Sec.~\ref{sec:discussion}.

\subsection{Evaluated Deblurring Methods}
\label{experiment: methods}
We evaluate nine state-of-the-art deblurring methods on the proposed MC-Blur dataset, including the multi-scale architectures (DeepDeblur \cite{nah2017deep}, SRN \cite{tao2018scale} and MIMO-UNet \cite{cho2021rethinking}), GAN based frameworks (DeblurGAN \cite{kupyn2018deblurgan}, DeblurGAN-v2 \cite{kupyn2019deblurgan}, and DBGAN \cite{zhang2020deblurring}), multi-patch networks (DMPHN \cite{Zhang_2019_CVPR} and MPRNet \cite{zamir2021multi}), and attention-based networks (Restormer \cite{zamir2021restormer}). We employ PSNR and SSIM as quantitative metrics to evaluate the deblurring methods while also comparing their performance qualitatively on synthesized and real blurry images. In addition, we use NIQE and SSEQ to evaluate the deblurring methods’ performance on real-world blur images.

\begin{table}[t] 
     \centering
      \caption{Performance evaluation of deep deblurring methods on the proposed UHDM set.}
        \begin{tabular}{c  c c } 
    \toprule
    \cellcolor{Gray} Method &  PSNR  & SSIM  \\
    \hline
    \cellcolor{Gray} DeepDeblur  \cite{nah2017deep} & 22.23 & 0.6322 \\
    \cellcolor{Gray} DeblurGAN \cite{kupyn2018deblurgan} & 20.39 & 0.5568 \\
    \cellcolor{Gray} SRN \cite{tao2018scale} & 22.28 & 0.6346  \\
    \cellcolor{Gray} DeblurGAN-v2 \cite{kupyn2019deblurgan} & 21.03 & 0.5839 \\
    \cellcolor{Gray} DMPHN \cite{Zhang_2019_CVPR} & 22.20 & 0.6378 \\
    \cellcolor{Gray} DBGAN \cite{zhang2020deblurring} & 21.52 & 0.6025 \\
    \cellcolor{Gray} MPRNet \cite{zamir2021multi} &  \textbf{23.70} & \textbf{0.7472}   \\
    \cellcolor{Gray} Restormer \cite{zamir2021restormer} & 22.39 & \underline{\textit{0.7356}} \\
    \cellcolor{Gray} MIMO-UNet \cite{cho2021rethinking} & \underline{\textit{22.97}} & 0.7317 \\
    \bottomrule
    \end{tabular}
    \label{table:results_conv}
\end{table}

\begin{table}[t] 
      \centering
        \caption{Performance evaluation of deep deblurring methods on the proposed LSD set.}
        \begin{tabular}{c  c c }
    \toprule
    \cellcolor{Gray} Method &  PSNR  & SSIM \\
    \hline
    \cellcolor{Gray} DeepDeblur  \cite{nah2017deep} & 20.73 & 0.7218   \\
    \cellcolor{Gray} DeblurGAN \cite{kupyn2018deblurgan} & 20.04 & 0.6335 \\
    \cellcolor{Gray} SRN \cite{tao2018scale} & 21.66 & 0.7664 \\
    \cellcolor{Gray} DeblurGAN-v2 \cite{kupyn2019deblurgan} & 21.13 & 0.6964 \\
    \cellcolor{Gray} DMPHN \cite{Zhang_2019_CVPR} & 21.23 & 0.7519 \\
    \cellcolor{Gray} DBGAN \cite{zhang2020deblurring} & 21.56 & 0.7536 \\
    \cellcolor{Gray} MPRNet \cite{zamir2021multi} & 21.32 & 0.7897 \\
    \cellcolor{Gray} Restormer \cite{zamir2021restormer} & \underline{\textit{22.35}} & \underline{\textit{0.8072}} \\
    \cellcolor{Gray} MIMO-UNet \cite{cho2021rethinking} & \textbf{22.56} & \textbf{0.8265} \\
    \bottomrule
    \end{tabular}
    \label{table:results_defocus}
\end{table}

\begin{table*}[t]
  \centering 
    \caption{Run-time and overhead comparison of SOTA deep deblurring methods. We adopt the number of parameters to measure the overhead of the model.}
   
    {
    \begin{tabular}{ccccc}
    \toprule
     \cellcolor{Gray} Method &  \cellcolor{Gray} DeepDeblur \cite{nah2017deep} & \cellcolor{Gray} DeblurGAN \cite{kupyn2018deblurgan} & \cellcolor{Gray} SRN \cite{tao2018scale} & \cellcolor{Gray} DeblurGAN-v2 \cite{kupyn2019deblurgan}   \\
    \hline 
        Speed (sec.) & 26.76 & \underline{\textit{2.46}} & 28.41 & 3.63    \\
         Params (M) & 11.72 & \textbf{6.07} & 6.88 & 7.84    \\
    \hline
    \cellcolor{Gray} DMPHN \cite{Zhang_2019_CVPR} & \cellcolor{Gray} DBGAN \cite{zhang2020deblurring} & \cellcolor{Gray} MPRNet \cite{zamir2021multi} & \cellcolor{Gray} Restormer \cite{zamir2021restormer} & \cellcolor{Gray} MIMO-UNet \cite{cho2021rethinking}  \\
    \hline 
         17.63 & 31.62 & 27.91 & 42.77 & \textbf{2.45}   \\
         21.69 & 11.59 & 20.13 & 26.10 & \underline{\textit{6.81}}  \\
    \bottomrule
    \end{tabular}}
    \label{table:speed}
\end{table*}%

\subsection{Benchmarking on the MC-Blur Dataset}
\label{experiment: mc}

\textbf{RHM Subset.} To evaluate image deblurring methods' performance, we first conduct experiments on RHM.
During the training stage, samples from all the training subsets of RHM are used together. We test them separately on the 250fps, 500fps, and 1000fps sets in the test stage.
Table~\ref{table:results_avg} shows that
DeepDeblur, SRN, DMPHN, MPRNet, Restormer, and MIMO-UNet perform well in terms of PSNR and SSIM. 
One contributing factor is the utilization of pixel-level loss functions in these methods, which significantly contributes to achieving elevated values in full-reference pixel-based metrics. 
The DeblurGAN, DeblurGAN-v2, and DBGAN models use discriminators to help synthesize more realistic deblurred images. 
These models are not only enforced to focus on pixel-wise measures (L1 or L2) but also pay attention to the whole image. Deblurring networks focusing on pixel-wise measures are expected to achieve higher values of PSNR and SSIM. MIMO-UNet and Restormer consistently outperform other methods, demonstrating the remarkable efficacy of their multi-scale schemes and transformer blocks. 
We also show a visual comparison of RHM in Fig.~\ref{fig:avg_1}.

\textbf{UHDM Subset.} In this part, the same deblurring methods are evaluated on the UHDM images. 
The training and testing samples are all from UHDM.
Table~\ref{table:results_conv} shows that all methods' PSNR and SSIM values are significantly lower than those in Table \ref{table:results_avg}. 
It could be attributed to several reasons. One reason is that we use large-size blur kernels to synthesize blurry images, making the UHDM set more difficult. 
Another reason is that compared with non-UHD image deblurring, deblurring UHD images require recovering more details. Therefore, UHD image deblurring is a more challenging task.
In addition, all the benchmarking deblurring networks are proposed for non-UHD image deblurring, which is also a reason causing their performance drop on UHD image deblurring.
Among these methods, MPRNet stands out as the top performer, underscoring the effectiveness of its multi-stage architecture. Furthermore, MIMO-UNet and Restormer also demonstrate strong performance in the UHD image deblurring setting.
We show qualitative results corresponding to the large-kernel blur caused in Fig. \ref{fig:conv_1}.

\begin{table*}[t]
  \centering 
    \caption{Performance comparison of cross-dataset learning regarding PSNR and SSIM, or NIQE and SSEQ. We train MPRNet \cite{zamir2021multi} on different training sets and evaluate the performance in different testing scenarios. Note that, for the RealBlur dataset \cite{rim2020real}, we employ the RealBlur-J subset. The top three and bottom two rows represent the results trained from a single source and multiple sources, respectively.}
    \setlength\tabcolsep{6.0pt}
    {
    \begin{tabular}{ c  c  c | c  c  c | c  c }
    \toprule
    \multicolumn{3}{c}{ \cellcolor{Gray}  Train}&\multicolumn{3}{c}{\cellcolor{Gray}  Test-1 (PSNR/SSIM$\uparrow$)}&\multicolumn{2}{c}{\cellcolor{Gray} Test-2 (NIQE/SSEQ$\downarrow$)}\\\cline{1-8}
    
   GoPro &  RealBlur-J & RHM & GoPro  & RealBlur-J & RHM & RMBQ & RWBI \\
    \hline
     \checkmark &  & & \textbf{30.05/0.9329} & 26.52/0.8635 & \underline{\textit{29.52/0.8914}} & \underline{\textit{6.0430/28.0522}} & \underline{\textit{5.6065/37.1958}} \\
      & \checkmark & & 23.45/0.8385 & \textbf{28.73/0.9011} & 24.67/0.8251 & 6.3704/34.9025 & 6.4757/40.2203 \\
      &  & \checkmark & \underline{\textit{30.04/0.9313}} & \underline{\textit{26.78/0.8732}} & \textbf{32.47/0.9297}  & \textbf{5.5654/25.5820} & \textbf{5.0310/35.0606} \\
      \hline
     \checkmark & \checkmark & & \underline{\textit{27.62/0.8970}} & \underline{\textit{28.71/0.8993}} & \underline{\textit{27.96/0.8632}}  & \underline{\textit{5.6983/29.7583}} & \underline{\textit{6.0287/38.8343}} \\
     \checkmark & \checkmark & \checkmark & \textbf{30.75/0.9395} & \textbf{29.80/0.9208} &  \textbf{32.63/0.9313} & \textbf{5.5039/26.6139} & \textbf{4.9222/34.3614} \\
    \bottomrule
    \end{tabular}}
    \label{tab:cross_dataset}
\end{table*}%

\begin{figure*}[t]
\begin{center}
 \includegraphics[width= \linewidth]{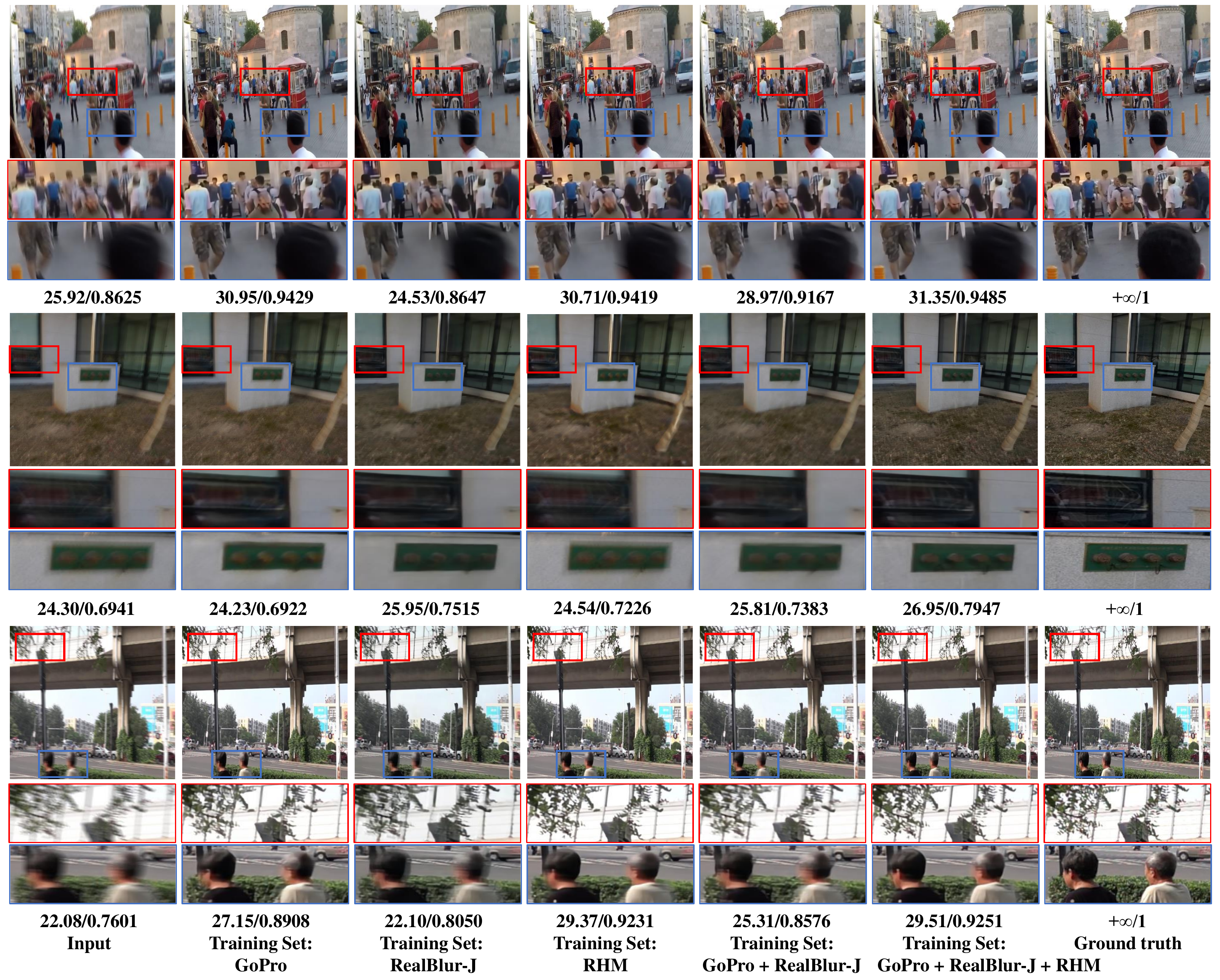} \small
   \caption{\textbf{Motion deblurred benchmarking results with cross-dataset learning}. The top, middle, and bottom rows show exemplary testing results from GoPro, RealBlur-J, and RHM. From left to right: input, results by identical deblurring network \cite{zamir2021multi} trained on GoPro, RealBlur-J, RHM, GoPro + RealBlur-J, and GoPro + RealBlur-J + RHM, and GT images. Zoom in for a better view.}
	 \label{fig:cross_dataset} 
\end{center}
\end{figure*}
\textbf{LSD Subset.} To investigate the performance of the SOTA deblurring methods in the case of defocus blur, we conduct a benchmark study on the LSD subset. %
All the training and test samples are from LSD.
Quantitative results are reported in Table~\ref{table:results_defocus}, and qualitative results are shown in Fig.~\ref{fig:defocus_1}. 
We note that defocus image deblurring is a more complex problem compared with deblurring of motion-blurred images. 
This is because defocus blur happens when the camera lens cannot converge all incoming light onto one sensor point, causing loss of vital image info, making it difficult to observe details and sharpen edges. Unlike motion blur, defocus severity is closely tied to object depth. Objects at varying distances experience different blurriness. This depth-dependent blur requires precise spatially varying kernel estimation for accurate deblurring.
While the SOTA deep deblurring methods can restore high-quality motion-deblurred images synthesized by averaging neighboring frames, the performance of defocus deblurring is significantly lower. 
Among these methods, MIMO-UNet, as a multi-scale deblurring approach, excels in performance, providing further evidence of the effectiveness of its multi-scale deblurring scheme for enhancing defocused image quality.

\textbf{RMBQ Subset.} In addition, we evaluate the performance of deblurring methods in real-world scenarios. 
The deblurred images by the evaluated methods (trained on RHM and tested on RMBQ) are shown in Fig.~\ref{fig:real_1} and Fig.~\ref{fig:real_2}. 
Overall, the deblurring methods trained on the proposed MC-Blur dataset generate sharp images given the input, recovering sharp text or image structure. Among these methods, MIMO-UNet exhibits superior qualitative performance in real-world blurry images. For instance, in Fig. \ref{fig:real_1}, the word `SPECIAL,' and in Fig. \ref{fig:real_2}, `WET PAINT,' appear sharper than all other methods. These results suggest that MIMO-UNet can outperform the benchmarking methods in real-world scenarios.

\subsection{Efficiency Analysis on UHD Images}
\label{exp:efficiency}

Efficiency should be considered when the image resolution is high, especially for UHD images. 
Table \ref{table:speed} shows the efficiency evaluation results on UHD images.
These experiments are carried out using a standard platform with a P40 GPU. 
The DeepDeblur \cite{nah2017deep}, SRN \cite{tao2018scale}, DMPHN \cite{Zhang_2019_CVPR}, DBGAN \cite{zhang2020deblurring}, MPRNet \cite{zamir2021multi} and Restormer \cite{zamir2021restormer} models require more than ten seconds to process one UHD image.
On the other hand, it takes $2.46$, $3.63$, $2.45$ seconds for DeblurGAN \cite{kupyn2018deblurgan}, DeblurGAN-v2 \cite{kupyn2019deblurgan} and MIMO-UNet \cite{cho2021rethinking} models, respectively, to process one UHD image.
We evaluate the speed and parameter efficiency of state-of-the-art deblurring methods. 
As such, we offer valuable insights into the efficiency of current methods in handling blurry images. Furthermore, this evaluation serves as a benchmark for the research community and underscores the practical applicability of our dataset.

\subsection{Cross-dataset Learning for Motion Deblurring}
\label{exp:dataset_comparison}

To illustrate the advantages of our RHM over the existing ones, we conduct an analysis with a cross-dataset learning protocol in this section. 
Specifically, we select a main-stream deblurring method \cite{zamir2021multi} and train this identical deblurring network using five combinations of different training sets. Then we evaluate the trained different versions of \cite{zamir2021multi} on various testing sets. Note that all parameter settings are set as the original paper, except that the epoch is set as $200$. Table \ref{tab:cross_dataset} shows the results.

Based on the top three rows in Table \ref{tab:cross_dataset}, the deblurring network is expected to achieve the best performance when the training and testing datasets come from the same source. It is also observed that our RHM provides better generalization potential to the deblurring network when trained with a single source than its counterparts. For instance, when tested on different test datasets, the network trained with RHM consistently achieves the best performance (see the \underline{\textit{italic}} values in the table). If the test set is not constrained to be different from the training source, the best performance will be achieved when the training and the test data sources are identical.

When trained with multiple sources (the bottom two rows in Table \ref{tab:cross_dataset}), the performance is significantly improved if our RHM is taken as an additional training source. Fig. \ref{fig:cross_dataset} demonstrates exemplar testing results corresponding to Test-1 in Table \ref{tab:cross_dataset} by the deblurring network \cite{zamir2021multi} trained with the five settings of training source. These qualitative results coincide with our analysis from Table \ref{tab:cross_dataset}.

In addition, we also evaluate on other real blurry datasets (\textit{i.e.}, the proposed RMBQ and RWBI datasets \cite{zhang2020deblurring}), which do not provide ground-truth sharp images. The results in Table \ref{tab:cross_dataset} demonstrate that the network trained with our RHM performs better in real-world scenery.

\begin{table}[t]
  \centering 
    \caption{Performance comparison of cross-dataset learning regarding PSNR and SSIM. We train MPRNet \cite{zamir2021multi} on different training sets and evaluate the performance in different testing scenarios. Though both LSD and DPDD consist of real defocus blurry images, a model trained on one dataset cannot achieve good performance on the other one.}
    \begin{tabular}{ c  c  c  c c }
    \toprule
    \multicolumn{2}{c}{ \cellcolor{Gray}  Train}&\multicolumn{3}{c}{\cellcolor{Gray}  Test (PSNR/SSIM$\uparrow$)} \\\cline{1-5}
    
   LSD &  DPDD & LSD & DPDD & LSD\&DPDD\\
    \hline
     \checkmark & & \textbf{24.58/0.8311} & \underline{\textit{20.77/0.7830}} & \textbf{{22.68/0.8071}} \\
      & \checkmark & \underline{\textit{23.77/0.8053}} & \textbf{21.32/0.7897} & \underline{\textit{22.55/0.7975}} \\
    \bottomrule
    \end{tabular}
    \label{tab:cross_dataset_defocus}
\end{table}%

\subsection{The Effectiveness of The LSD Set}
\label{exp:dataset_comparison_defocus}

As mentioned above, there exists a high-quality real defocus blurry DPDD dataset. Compared to the DPDD dataset, the proposed LSD set consists of heavier defocus blurry images; thus, there is a significant difference between them. In this section, to verify the effectiveness of the proposed LSD, we conduct another experimental study.

Similarly, we train an identical deblurring method \cite{zamir2021multi}, using our proposed LSD and DPDD respectively as training sets and then evaluate the trained different versions on different testing sets, including LSD, DPDD and the joint of LSD and DPDD.
Table \ref{tab:cross_dataset_defocus} shows that, though both LSD and DPDD consist of real defocus blurry images, a model trained on one dataset cannot perform well on the other. 
Specifically, the model trained on LSD performs better for heavy-defocus deblurring, while the model trained on DPDD performs better on non-heavy-defocus blurry images. It is also important to observe that, for the testing set consisting of the joint of LSD and DPDD, the model trained with LSD performs better than the one prepared with DPDD. This suggests the utility advance of LSD over DPDD. We attribute this to the heavy defocus blurry images in the LSD set.

\subsection{Cross-set Learning Regarding Blur Factors}
\label{exp:dataset_comparison_different_blur}

To further study how different blur factors influence the behavior of SOTA methods, we examine how methods trained on the motion blur subset (\textit{i.e.}, RHM) behave on subsets of other factors (\textit{i.e.}, LSD and UHDM). We train MPRNet, Restormer, and MIMO-UNet on the RHM and test these methods respectively on UHDM and LSD.

Experimental results in Table~\ref{table:cross-sets_2} show that models trained on one type of blurry images perform poorly on recovering sharp images from other types of blur. Therefore, proposing a method that can recover sharp images from various kinds of blurry images is still an open topic.

\begin{table}[t] 
     \centering
      \caption{Performance of representative deep deblurring methods trained on the RHM set and evaluated on the UHDM and LSD subsets.}
        \begin{tabular}{ccc} 
    \toprule
    \cellcolor{Gray} Method &  UHDM (PSNR/SSIM)  & LSD (PSNR/SSIM)  \\
    \hline
    \cellcolor{Gray} MPRNet \cite{zamir2021multi} & \underline{\textit{21.18}}/0.7012  &  19.97/0.7784  \\
    \cellcolor{Gray} Restormer \cite{zamir2021restormer} & 20.89/\underline{\textit{0.7048}} & \underline{\textit{19.98/0.7785}} \\
    \cellcolor{Gray} MIMO-UNet \cite{cho2021rethinking} & \textbf{21.24/0.7051} & \textbf{20.00/0.7806} \\
    \bottomrule
    \end{tabular}
    \label{table:cross-}
\end{table}

\subsection{Discussion}
\label{sec:discussion}

The benchmarking results on different sets of the proposed MC-Blur dataset, reveal several interesting findings. 
First, GAN-based networks achieve lower values of PSNR \& SSIM for motion-blurred images than methods without using the GAN framework. However, the two networks (with and without using the GAN framework) show fewer differences for defocus images. This indicates that paying attention to the whole images (\textit{e.g.}, the adversarial loss function), rather than just considering the pixel level (\textit{e.g.}, L1 and L2 loss functions), may be a direction for defocus deblurring. The results in Tables \ref{table:results_avg}, \ref{table:results_defocus} support this finding.
Second, current deep deblurring networks can generate high-quality images for motion-blurred images. However, it is difficult for them to achieve similar performance on large kernel-based Ultra-High-Definition blurry images (Tables \ref{table:results_avg}, \ref{table:results_conv}). Since increasing numbers of modern mobile devices allow capturing UHD images, it may be a meaningful direction for researchers to study UHD image deblurring. 
Third, current deep methods can deblur a non-UHD image in two seconds \cite{Zhang_2019_CVPR}. However, as shown in Table \ref{table:speed}, handling a UHD image takes significantly longer. Therefore, generating deblurred UHD images at a high rate while maintaining deblurring performance is still an open problem.
In addition, this work has limitations. First, the proposed dataset is mainly for benchmarking single-image deblurring; thus, we do not evaluate video deblurring methods. Second, the MC-Blur dataset does not include human faces, so it is unsuitable for face restoration. Third, except the RMBQ subset, all the other subsets focus on a single blur factor.

\section{Conclusion}
\label{Conclusion}
We establish the first large-scale multi-cause image deblurring dataset, MC-Blur, to benchmark deblurring methods on images with blur caused by various factors.
The MC-Blur dataset includes a real high fps-based motion-blurred set, a large-kernel Ultra-High-Definition motion-blurred set, a large-scale heavy defocus blurry set and a real mixed blurry set. 
Based on these unique sets of images, the current SOTA image deblurring approaches are benchmarked to study their advances and limitations in diverse scenarios. Cross-dataset benchmarking is also carried out to verify the advantage of the proposed MC-Blur dataset. As such, we supply a comprehensive understanding of the SOTA image deblurring methods. 
The established MC-Blur is expected to drive the community's research of multi-cause image deblurring. 
In the future, we plan to expand our efforts by generating additional datasets for evaluating video deblurring methods, assessing the performance of current techniques on images afflicted by more than two degrading factors, and exploring the potential advantages of deblurring in applications such as video segmentation and tracking.

\bibliographystyle{IEEEtran}
\bibliography{egbib}

\begin{thebibliography}{10}
\providecommand{\url}[1]{#1}
\csname url@samestyle\endcsname
\providecommand{\newblock}{\relax}
\providecommand{\bibinfo}[2]{#2}
\providecommand{\BIBentrySTDinterwordspacing}{\spaceskip=0pt\relax}
\providecommand{\BIBentryALTinterwordstretchfactor}{4}
\providecommand{\BIBentryALTinterwordspacing}{\spaceskip=\fontdimen2\font plus
\BIBentryALTinterwordstretchfactor\fontdimen3\font minus \fontdimen4\font\relax}
\providecommand{\BIBforeignlanguage}[2]{{%
\expandafter\ifx\csname l@#1\endcsname\relax
\typeout{** WARNING: IEEEtran.bst: No hyphenation pattern has been}%
\typeout{** loaded for the language `#1'. Using the pattern for}%
\typeout{** the default language instead.}%
\else
\language=\csname l@#1\endcsname
\fi
#2}}
\providecommand{\BIBdecl}{\relax}
\BIBdecl

\bibitem{zhang2022deep}
K.~Zhang, W.~Ren, W.~Luo, W.-S. Lai, B.~Stenger, M.-H. Yang, and H.~Li, ``Deep image deblurring: A survey,'' \emph{International Journal of Computer Vision}, vol. 130, no.~9, pp. 2103--2130, 2022.

\bibitem{Nah_2019_CVPR_Workshops_REDS}
S.~Nah, S.~Baik, S.~Hong, G.~Moon, S.~Son, R.~Timofte, and K.~M. Lee, ``Ntire 2019 challenge on video deblurring and super-resolution: Dataset and study,'' in \emph{IEEE Conference on Computer Vision and Pattern Recognition Workshop}, 2019.

\bibitem{kohler2012recording}
R.~K{\"o}hler, M.~Hirsch, B.~Mohler, B.~Sch{\"o}lkopf, and S.~Harmeling, ``Recording and playback of camera shake: Benchmarking blind deconvolution with a real-world database,'' in \emph{European Conference on Computer Vision}, 2012.

\bibitem{rim2020real}
J.~Rim, H.~Lee, J.~Won, and S.~Cho, ``Real-world blur dataset for learning and benchmarking deblurring algorithms,'' in \emph{European Conference on Computer Vision}, 2020.

\bibitem{abuolaim2020defocus}
A.~Abuolaim and M.~S. Brown, ``Defocus deblurring using dual-pixel data,'' in \emph{European Conference on Computer Vision}, 2020.

\bibitem{levin2009understanding}
A.~Levin, Y.~Weiss, F.~Durand, and W.~T. Freeman, ``Understanding and evaluating blind deconvolution algorithms,'' in \emph{IEEE Conference on Computer Vision and Pattern Recognition}, 2009.

\bibitem{sun2012super}
L.~Sun and J.~Hays, ``Super-resolution from internet-scale scene matching,'' in \emph{IEEE International Conference on Computational Photography}, 2012.

\bibitem{lai2016comparative}
W.-S. Lai, J.-B. Huang, Z.~Hu, N.~Ahuja, and M.-H. Yang, ``A comparative study for single image blind deblurring,'' in \emph{IEEE Conference on Computer Vision and Pattern Recognition}, 2016.

\bibitem{nah2017deep}
S.~Nah, T.~Hyun~Kim, and K.~Mu~Lee, ``Deep multi-scale convolutional neural network for dynamic scene deblurring,'' in \emph{IEEE Conference on Computer Vision and Pattern Recognition}, 2017.

\bibitem{shen2019human}
Z.~Shen, W.~Wang, X.~Lu, J.~Shen, H.~Ling, T.~Xu, and L.~Shao, ``Human-aware motion deblurring,'' in \emph{IEEE International Conference on Computer Vision}, 2019.

\bibitem{jiang2020learning}
Z.~Jiang, Y.~Zhang, D.~Zou, J.~Ren, J.~Lv, and Y.~Liu, ``Learning event-based motion deblurring,'' in \emph{IEEE Conference on Computer Vision and Pattern Recognition}, 2020.

\bibitem{su2017deep}
S.~Su, M.~Delbracio, J.~Wang, G.~Sapiro, W.~Heidrich, and O.~Wang, ``Deep video deblurring for hand-held cameras,'' in \emph{IEEE Conference on Computer Vision and Pattern Recognition}, 2017.

\bibitem{hradivs2015convolutional}
M.~Hradi{\v{s}}, J.~Kotera, P.~Zemc{\i}k, and F.~{\v{S}}roubek, ``Convolutional neural networks for direct text deblurring,'' in \emph{British Machine Vision Conference}, 2015.

\bibitem{shen2018deep}
Z.~Shen, W.-S. Lai, T.~Xu, J.~Kautz, and M.-H. Yang, ``Deep semantic face deblurring,'' in \emph{IEEE Conference on Computer Vision and Pattern Recognition}, 2018.

\bibitem{zhou2019davanet}
S.~Zhou, J.~Zhang, W.~Zuo, H.~Xie, J.~Pan, and J.~S. Ren, ``Davanet: Stereo deblurring with view aggregation,'' in \emph{IEEE Conference on Computer Vision and Pattern Recognition}, 2019.

\bibitem{hirsch2011fast}
M.~Hirsch, C.~J. Schuler, S.~Harmeling, and B.~Sch{\"o}lkopf, ``Fast removal of non-uniform camera shake,'' in \emph{IEEE International Conference on Computer Vision}, 2011.

\bibitem{bai2019single}
Y.~Bai, H.~Jia, M.~Jiang, X.~Liu, X.~Xie, and W.~Gao, ``Single-image blind deblurring using multi-scale latent structure prior,'' \emph{IEEE Transactions on Circuits and Systems for Video Technology}, vol.~30, no.~7, pp. 2033--2045, 2019.

\bibitem{wen2020simple}
F.~Wen, R.~Ying, Y.~Liu, P.~Liu, and T.-K. Truong, ``A simple local minimal intensity prior and an improved algorithm for blind image deblurring,'' \emph{IEEE Transactions on Circuits and Systems for Video Technology}, vol.~31, no.~8, pp. 2923--2937, 2020.

\bibitem{luo2021blind}
B.~Luo, Z.~Cheng, L.~Xu, G.~Zhang, and H.~Li, ``Blind image deblurring via superpixel segmentation prior,'' \emph{IEEE Transactions on Circuits and Systems for Video Technology}, vol.~32, no.~3, pp. 1467--1482, 2021.

\bibitem{pan2014deblurring}
J.~Pan, Z.~Hu, Z.~Su, and M.-H. Yang, ``Deblurring text images via l0-regularized intensity and gradient prior,'' in \emph{IEEE Conference on Computer Vision and Pattern Recognition}, 2014.

\bibitem{pan2017deblurring}
J.~Pan, D.~Sun, H.~Pfister, and M.-H. Yang, ``Deblurring images via dark channel prior,'' \emph{IEEE Transactions on Pattern Analysis and Machine Intelligence}, vol.~40, no.~10, pp. 2315--2328, 2017.

\bibitem{chen2019blind}
L.~Chen, F.~Fang, T.~Wang, and G.~Zhang, ``Blind image deblurring with local maximum gradient prior,'' in \emph{IEEE Conference on Computer Vision and Pattern Recognition}, 2019.

\bibitem{wang2023self}
S.~Wang, C.~Li, Y.~Li, Y.~Yuan, and G.~Wang, ``Self-supervised information bottleneck for deep multi-view subspace clustering,'' \emph{IEEE Transactions on Image Processing}, vol.~32, pp. 1555--1567, 2023.

\bibitem{li2018joint}
C.~Li, X.~Wang, W.~Dong, J.~Yan, Q.~Liu, and H.~Zha, ``Joint active learning with feature selection via cur matrix decomposition,'' \emph{IEEE transactions on pattern analysis and machine intelligence}, vol.~41, no.~6, pp. 1382--1396, 2018.

\bibitem{li2018dynamic}
C.~Li, F.~Wei, W.~Dong, X.~Wang, Q.~Liu, and X.~Zhang, ``Dynamic structure embedded online multiple-output regression for streaming data,'' \emph{IEEE transactions on pattern analysis and machine intelligence}, vol.~41, no.~2, pp. 323--336, 2018.

\bibitem{sun2015learning}
J.~Sun, W.~Cao, Z.~Xu, and J.~Ponce, ``Learning a convolutional neural network for non-uniform motion blur removal,'' in \emph{IEEE Conference on Computer Vision and Pattern Recognition}, 2015.

\bibitem{chakrabarti2016neural}
A.~Chakrabarti, ``A neural approach to blind motion deblurring,'' in \emph{European Conference on Computer Vision}, 2016.

\bibitem{nimisha2017blur}
T.~M. Nimisha, A.~Kumar~Singh, and A.~N. Rajagopalan, ``Blur-invariant deep learning for blind-deblurring,'' in \emph{IEEE International Conference on Computer Vision}, 2017.

\bibitem{xu2017motion}
X.~Xu, J.~Pan, Y.-J. Zhang, and M.-H. Yang, ``Motion blur kernel estimation via deep learning,'' \emph{IEEE Transactions on Image Processing}, vol.~27, no.~1, pp. 194--205, 2017.

\bibitem{jin2018learning}
M.~Jin, G.~Meishvili, and P.~Favaro, ``Learning to extract a video sequence from a single motion-blurred image,'' in \emph{IEEE Conference on Computer Vision and Pattern Recognition}, 2018.

\bibitem{hyun2017online}
T.~Hyun~Kim, K.~Mu~Lee, B.~Scholkopf, and M.~Hirsch, ``Online video deblurring via dynamic temporal blending network,'' in \emph{IEEE International Conference on Computer Vision}, 2017.

\bibitem{aittala2018burst}
M.~Aittala and F.~Durand, ``Burst image deblurring using permutation invariant convolutional neural networks,'' in \emph{European Conference on Computer Vision}, 2018.

\bibitem{nah2019recurrent}
S.~Nah, S.~Son, and K.~M. Lee, ``Recurrent neural networks with intra-frame iterations for video deblurring,'' in \emph{IEEE Conference on Computer Vision and Pattern Recognition}, 2019.

\bibitem{wang2019edvr}
X.~Wang, K.~C. Chan, K.~Yu, C.~Dong, and C.~Change~Loy, ``{EDVR}: Video restoration with enhanced deformable convolutional networks,'' in \emph{IEEE Conference on Computer Vision and Pattern Recognition Workshop}, 2019.

\bibitem{mustaniemi2019gyroscope}
J.~Mustaniemi, J.~Kannala, S.~S{\"a}rkk{\"a}, J.~Matas, and J.~Heikkila, ``Gyroscope-aided motion deblurring with deep networks,'' in \emph{IEEE Winter Conference on Applications of Computer Vision}, 2019.

\bibitem{aljadaany2019douglas}
R.~Aljadaany, D.~K. Pal, and M.~Savvides, ``Douglas-rachford networks: Learning both the image prior and data fidelity terms for blind image deconvolution,'' in \emph{IEEE Conference on Computer Vision and Pattern Recognition}, 2019.

\bibitem{kaufman2020deblurring}
A.~Kaufman and R.~Fattal, ``Deblurring using analysis-synthesis networks pair,'' in \emph{IEEE Conference on Computer Vision and Pattern Recognition}, 2020.

\bibitem{kupyn2018deblurgan}
O.~Kupyn, V.~Budzan, M.~Mykhailych, D.~Mishkin, and J.~Matas, ``Deblurgan: Blind motion deblurring using conditional adversarial networks,'' in \emph{IEEE Conference on Computer Vision and Pattern Recognition}, 2018.

\bibitem{kupyn2019deblurgan}
O.~Kupyn, T.~Martyniuk, J.~Wu, and Z.~Wang, ``Deblurgan-v2: Deblurring (orders-of-magnitude) faster and better,'' in \emph{IEEE International Conference on Computer Vision}, 2019.

\bibitem{Zhang_2019_CVPR}
H.~Zhang, Y.~Dai, H.~Li, and P.~Koniusz, ``Deep stacked hierarchical multi-patch network for image deblurring,'' in \emph{IEEE Conference on Computer Vision and Pattern Recognition}, 2019.

\bibitem{zhang2018dynamic}
J.~Zhang, J.~Pan, J.~Ren, Y.~Song, L.~Bao, R.~W. Lau, and M.-H. Yang, ``Dynamic scene deblurring using spatially variant recurrent neural networks,'' in \emph{IEEE Conference on Computer Vision and Pattern Recognition}, 2018.

\bibitem{tao2018scale}
X.~Tao, H.~Gao, X.~Shen, J.~Wang, and J.~Jia, ``Scale-recurrent network for deep image deblurring,'' in \emph{IEEE Conference on Computer Vision and Pattern Recognition}, 2018.

\bibitem{zamir2021multi}
S.~W. Zamir, A.~Arora, S.~Khan, M.~Hayat, F.~S. Khan, M.-H. Yang, and L.~Shao, ``Multi-stage progressive image restoration,'' in \emph{IEEE Conference on Computer Vision and Pattern Recognition}, 2021.

\bibitem{zamir2021restormer}
S.~W. Zamir, A.~Arora, S.~Khan, M.~Hayat, F.~S. Khan, and M.-H. Yang, ``Restormer: Efficient transformer for high-resolution image restoration,'' in \emph{IEEE Conference on Computer Vision and Pattern Recognition}, 2022.

\bibitem{gao2019dynamic}
H.~Gao, X.~Tao, X.~Shen, and J.~Jia, ``Dynamic scene deblurring with parameter selective sharing and nested skip connections,'' in \emph{IEEE Conference on Computer Vision and Pattern Recognition}, 2019.

\bibitem{ren2019neural}
D.~Ren, K.~Zhang, Q.~Wang, Q.~Hu, and W.~Zuo, ``Neural blind deconvolution using deep priors,'' in \emph{IEEE Conference on Computer Vision and Pattern Recognition}, 2020.

\bibitem{zhou2019spatio}
S.~Zhou, J.~Zhang, J.~Pan, H.~Xie, W.~Zuo, and J.~Ren, ``Spatio-temporal filter adaptive network for video deblurring,'' in \emph{IEEE International Conference on Computer Vision}, 2019.

\bibitem{pan2020cascaded}
J.~Pan, H.~Bai, and J.~Tang, ``Cascaded deep video deblurring using temporal sharpness prior,'' in \emph{IEEE Conference on Computer Vision and Pattern Recognition}, 2020.

\bibitem{zhang2018adversarial}
K.~Zhang, W.~Luo, Y.~Zhong, L.~Ma, W.~Liu, and H.~Li, ``Adversarial spatio-temporal learning for video deblurring,'' \emph{IEEE Transactions on Image Processing}, vol.~28, no.~1, pp. 291--301, 2018.

\bibitem{niu2021blind}
W.~Niu, K.~Zhang, W.~Luo, and Y.~Zhong, ``Blind motion deblurring super-resolution: When dynamic spatio-temporal learning meets static image understanding,'' \emph{IEEE Transactions on Image Processing}, vol.~30, pp. 7101--7111, 2021.

\bibitem{niu2021deep}
W.~Niu, K.~Zhang, W.~Luo, Y.~Zhong, and H.~Li, ``Deep robust image deblurring via blur distilling and information comparison in latent space,'' \emph{Neurocomputing}, vol. 466, pp. 69--79, 2021.

\bibitem{zhang2020deblurring}
K.~Zhang, W.~Luo, Y.~Zhong, B.~Stenger, L.~Ma, W.~Liu, and H.~Li, ``Deblurring by realistic blurring,'' in \emph{IEEE Conference on Computer Vision and Pattern Recognition}, 2020.

\bibitem{chen2018reblur2deblur}
H.~Chen, J.~Gu, O.~Gallo, M.-Y. Liu, A.~Veeraraghavan, and J.~Kautz, ``Reblur2deblur: Deblurring videos via self-supervised learning,'' in \emph{IEEE International Conference on Computational Photography}, 2018.

\bibitem{rim2022realistic}
J.~Rim, G.~Kim, J.~Kim, J.~Lee, S.~Lee, and S.~Cho, ``Realistic blur synthesis for learning image deblurring,'' in \emph{European Conference on Computer Vision}, 2022.

\bibitem{madam2018unsupervised}
T.~Madam~Nimisha, K.~Sunil, and A.~Rajagopalan, ``Unsupervised class-specific deblurring,'' in \emph{European Conference on Computer Vision}, 2018.

\bibitem{lu2019unsupervised}
B.~Lu, J.-C. Chen, and R.~Chellappa, ``Unsupervised domain-specific deblurring via disentangled representations,'' in \emph{IEEE Conference on Computer Vision and Pattern Recognition}, 2019.

\bibitem{li2023self}
J.~Li, W.~Wang, Y.~Nan, and H.~Ji, ``Self-supervised blind motion deblurring with deep expectation maximization,'' in \emph{IEEE Conference on Computer Vision and Pattern Recognition}, 2023.

\bibitem{purohit2019region}
K.~Purohit and A.~Rajagopalan, ``Region-adaptive dense network for efficient motion deblurring,'' in \emph{Proceedings of the AAAI Conference on Artificial Intelligence}, vol.~34, no.~07, 2020, pp. 11\,882--11\,889.

\bibitem{suin2020spatially}
M.~Suin, K.~Purohit, and A.~Rajagopalan, ``Spatially-attentive patch-hierarchical network for adaptive motion deblurring,'' in \emph{IEEE Conference on Computer Vision and Pattern Recognition}, 2020.

\bibitem{park2020multi}
P.~D. Park, D.~U. Kang, J.~Kim, and S.~Y. Chun, ``Multi-temporal recurrent neural networks for progressive non-uniform single image deblurring with incremental temporal training,'' in \emph{European Conference on Computer Vision}, 2020.

\bibitem{zhang2021multi}
X.~Zhang, T.~Wang, R.~Jiang, L.~Zhao, and Y.~Xu, ``Multi-attention convolutional neural network for video deblurring,'' \emph{IEEE Transactions on Circuits and Systems for Video Technology}, vol.~32, no.~4, pp. 1986--1997, 2021.

\bibitem{sun2022event}
L.~Sun, C.~Sakaridis, J.~Liang, Q.~Jiang, K.~Yang, P.~Sun, Y.~Ye, K.~Wang, and L.~V. Gool, ``Event-based fusion for motion deblurring with cross-modal attention,'' in \emph{European Conference on Computer Vision}, 2022.

\bibitem{kong2023efficient}
L.~Kong, J.~Dong, J.~Ge, M.~Li, and J.~Pan, ``Efficient frequency domain-based transformers for high-quality image deblurring,'' in \emph{IEEE Conference on Computer Vision and Pattern Recognition}, 2023.

\bibitem{dudhane2023burstormer}
A.~Dudhane, S.~W. Zamir, S.~Khan, F.~S. Khan, and M.-H. Yang, ``Burstormer: Burst image restoration and enhancement transformer,'' \emph{IEEE Conference on Computer Vision and Pattern Recognition}, 2023.

\bibitem{park2023all}
D.~Park, B.~H. Lee, and S.~Y. Chun, ``All-in-one image restoration for unknown degradations using adaptive discriminative filters for specific degradations,'' in \emph{IEEE Conference on Computer Vision and Pattern Recognition}.\hskip 1em plus 0.5em minus 0.4em\relax IEEE, 2023.

\bibitem{li2021arvo}
D.~Li, C.~Xu, K.~Zhang, X.~Yu, Y.~Zhong, W.~Ren, H.~Suominen, and H.~Li, ``Arvo: Learning all-range volumetric correspondence for video deblurring,'' in \emph{Proceedings of the IEEE/CVF Conference on Computer Vision and Pattern Recognition}, 2021, pp. 7721--7731.

\bibitem{hu2021pyramid}
X.~Hu, W.~Ren, K.~Yu, K.~Zhang, X.~Cao, W.~Liu, and B.~Menze, ``Pyramid architecture search for real-time image deblurring,'' in \emph{IEEE International Conference on Computer Vision}, 2021.

\bibitem{wang2023bad}
P.~Wang, L.~Zhao, R.~Ma, and P.~Liu, ``Bad-nerf: Bundle adjusted deblur neural radiance fields,'' in \emph{IEEE Conference on Computer Vision and Pattern Recognition}, 2023.

\bibitem{boracchi2012modeling}
G.~Boracchi and A.~Foi, ``Modeling the performance of image restoration from motion blur,'' \emph{IEEE Transactions on Image Processing}, 2012.

\bibitem{cho2021rethinking}
S.-J. Cho, S.-W. Ji, J.-P. Hong, S.-W. Jung, and S.-J. Ko, ``Rethinking coarse-to-fine approach in single image deblurring,'' in \emph{IEEE International Conference on Computer Vision}, 2021.

\end{thebibliography}

\end{document}